\documentclass[nonacm,natbib=false,sigconf]{acmart}
\usepackage{graphicx}
\usepackage{caption,subcaption}
\usepackage[para]{footmisc}
\usepackage{booktabs}
\usepackage{multirow}
\usepackage{xcolor,colortbl}
\usepackage{hyperref}
\usepackage{cleveref}
\usepackage[justification=centering]{caption}

\usepackage[style=ACM-Reference-Format,backend=bibtex,sorting=none]{biblatex}
\begin{document}

\title{DeepHateExplainer: Explainable Hate Speech Detection in Under-resourced Bengali Language}


\author{Md. Rezaul Karim}
\affiliation{%
  \institution{Fraunhofer FIT \& RWTH Aachen University, Aachen, Germany}
   \country{}
   }

\author{Sumon Kanti Dey}
\affiliation{%
  \institution{Noakhali Science and Technology University, Bangladesh}
  \country{}}
  
\author{Tanhim Islam}
\affiliation{
   \institution{RWTH Aachen University}
   \city{Aachen}
   \country{Germany}}
   
\author{Sagor Sarker}
\affiliation{
   \institution{Begum Rokeya University}
   \city{Rangpur}
   \country{Bangladesh}}
   
\author{Mehadi Hasan Menon}
\affiliation{
   \institution{Begum Rokeya University}
   \city{Rangpur}
   \country{Bangladesh}}
   
\author{Kabir Hossain}
\affiliation{
   \institution{The University of Alabama}
   \city{Tuscaloosa}
   \country{USA}}

\author{Bharathi Raja Chakravarthi}
\affiliation{%
  \institution{National University of Ireland, Galway, Ireland}
  \country{}}  

\author{Md. Azam Hossain}
\affiliation{%
  \institution{Islamic University of Technology}
  \city{Gazipur}
  \country{Bangladesh}}  
  
\author{Stefan Decker}
\affiliation{%
  \institution{Fraunhofer FIT \& RWTH Aachen University, Aachen, Germany}
   \country{}
   }

\renewcommand{\shortauthors}{Karim, Sumon, and Decker et al.}

\begin{abstract}
  The exponential growths of social media and micro-blogging sites not only provide platforms for empowering freedom of expressions and individual voices, but also enables people to express anti-social behavior like online harassment, cyberbullying, and hate speech. Numerous works have been proposed to utilize textual data for social and anti-social behavior analysis, by predicting the contexts mostly for highly-resourced languages like English. However, some languages are under-resourced, e.g., South Asian languages like Bengali, that lack computational resources for accurate natural language processing~(NLP). In this paper\footnote{Proceeding of IEEE International Conference on Data Science and Advanced Analytics~(DSAA'2021), October 6-9, 2021, Porto, Portugal.}, we propose an explainable approach for hate speech detection from the under-resourced Bengali language, which we called \emph{DeepHateExplainer}. Bengali texts are first comprehensively preprocessed, before classifying them into political, personal, geopolitical, and religious hates using a neural ensemble method of transformer-based neural architectures~(i.e., monolingual Bangla BERT-base, multilingual BERT-cased/uncased, and XLM-RoBERTa). Important~(most and least) terms are then identified using sensitivity analysis and layer-wise relevance propagation~(LRP), before providing human-interpretable explanations\footnote{To foster reproducible research, we make available the data, source codes, models, and notebooks: \textcolor{blue}{\url{https://github.com/rezacsedu/DeepHateExplainer}}} for the hate speech detection. Finally, we compute comprehensiveness and sufficiency scores to measure the quality of explanations w.r.t faithfulness. Evaluations against machine learning~(linear and tree-based models) and neural networks~(i.e., CNN, Bi-LSTM, and Conv-LSTM with word embeddings) baselines yield F1-scores of 78\%, 91\%, 89\%, and 84\%, for political, personal, geopolitical, and religious hates, respectively, outperforming both ML and DNN baselines. 
\end{abstract}
\maketitle

\section{Introduction}
Exponential growths of micro-blogging sites and social media not only empower freedom of expressions and individual voices, but also enables people to express anti-social behavior~\cite{elsherief2018hate,karim2020classification}, such as cyberbullying, online rumours, and spreading hatred statements~\cite{ribeiro2018characterizing,karim2020classification}.
Abusive speech expressing prejudice towards a certain group is also very common~\cite{karim2020classification}, and based on race, religion, and sexual orientation is getting pervasive.
United Nations Strategy and Plan of Action on Hate Speech~\cite{guterres2019united} defines hate speech as \emph{any kind of communication in speech, writing or behaviour, that attacks or uses pejorative or discriminatory language regarding a person or a group based on their religion, ethnicity, colour, gender or other identity factors}.
Bengali is spoken by 230 million people in Bangladesh and India~\cite{islam2009research}, making it one of the major languages in the world. Although Bengali is a rich language with a lot of diversity, it is severely low-resourced for natural language processing~(NLP). This is mainly due to the lack of necessary computational resources such as language models, labelled datasets, and efficient machine learning~(ML) methods for various NLP tasks. Similar to other major languages like English, the use of hate speech in Bengali is also getting rampant, which is due to unrestricted access and use of social media and digitalization~\cite{zhang2018detecting}.

\begin{figure*}
    \centering
    \includegraphics[scale=0.92]{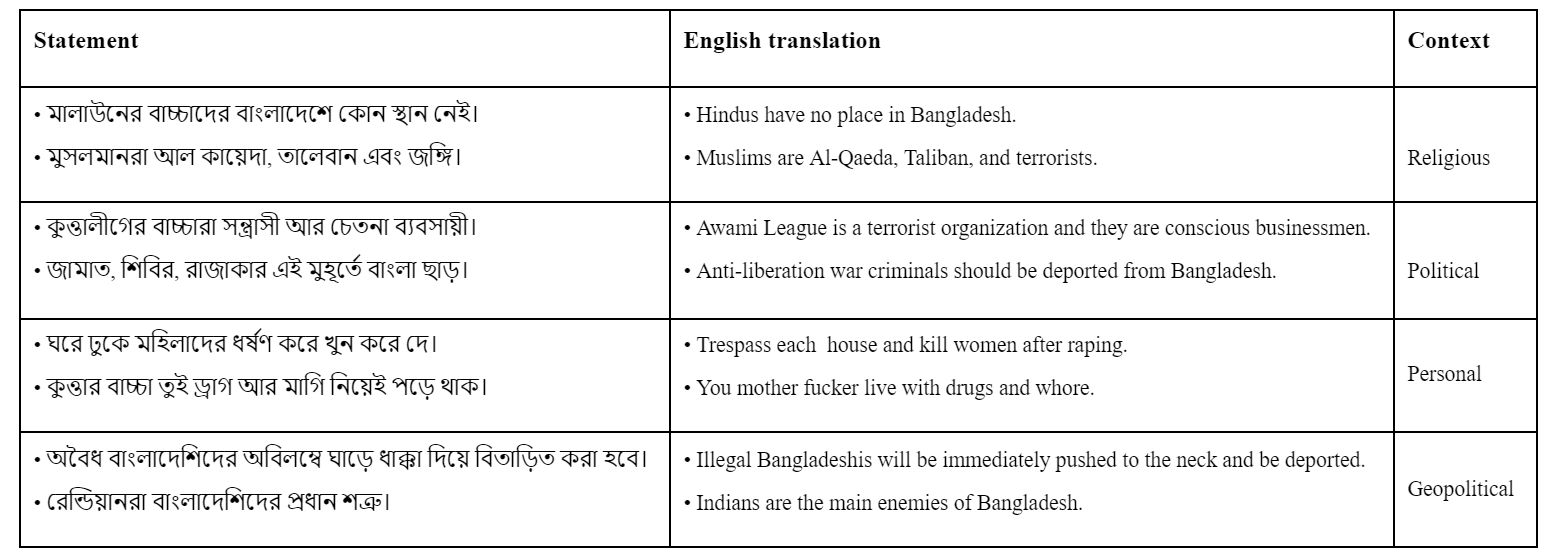}
    \caption{Example hate speech in Bengali, either directed towards a specific person or entity, or generalized towards a group}    
    \label{cdec_wf3}
\end{figure*}

Some examples of Bengali hate speech and their respective English translations are shown in Fig. \ref{cdec_wf3} that are either directed towards a specific person or entity or generalized towards a group. These examples signify how severe Bengali hateful statements could be. Nevertheless, there is a potential chance that these could lead to serious consequences such as hate crimes~\cite{karim2020classification}, regardless of languages, geographic locations, or ethnicity. Automatic identification of hate speech and raising public awareness is a non-trivial task~\cite{karim2020classification}. However, manually reviewing and verifying a large volume of online content is not only time-consuming but also labor-intensive~\cite{hate5}. Further, accurate identification requires automated and robust ML methods. Compared to traditional ML and neural networks~(DNNs)-based approaches, state-of-the-art~(SotA) language models are becoming increasingly effective. Nevertheless, a serious drawback of many existing approaches is that the outputs can neither be traced back to the inputs, nor it is clear why outputs are transformed in a certain way. This makes even the most efficient language models \emph{black-box} methods. 
Therefore, how a prediction is made by an algorithm should be as transparent as possible to users to gain human trust in AI systems.  

To mitigate the opaqueness of \emph{black-box} models and inspired by recent successes of transformer language models~(e.g., BERT~\cite{devlin2018BERT}, RoBERTa~\cite{liu2019roberta}, XLNet~\cite{yang2019xlnet}, and ELECTRA~\cite{clark2020electra}), we propose \emph{DeepHateExplainer} - an explainable approach for hate speech detection from \emph{under-resourced} Bengali language. Our approach is based on ensemble of BERT variants, including monolingual Bangla BERT-base~\cite{Sagor_2020}, m-BERT~(cased/uncased), and XLM-RoBERTa. Further, we provide global and local explanations of the predictions in a post-hoc fashion and measures of explanations w.r.t faithfulness.

\section{Related Work}\label{rw}
Numerous works have been proposed to accurately and reliably identification of hate speech from major languages like English~\cite{elsherief2018hate,hate5}. Classic methods traditionally rely on manual feature engineering, e.g., support vector machines~(SVM), Na{\"i}ve Bayes~(NB), logistic regression~(LR), decision trees~(DT), random forest~(RF), and gradient boosted trees~(GBT). On the other hand, DNN-based approaches that learn multilayers of abstract features from raw texts, are primarily based on convolutional~(CNN) or long short-term memory~(LSTM) networks. In comparison with DNNs, these approaches are rather incomparable as the efficiency of linear models at dealing with billions of such texts proven less accurate and unscalable. 
CNN and LSTM are two popular DNN architectures: CNN is an effective feature extractor, whereas LSTM is suitable for modelling orderly sequence learning problems. CNN extracts word or character combinations, e.g., n-grams, and LSTM learns long-range word or character dependencies in texts.
While each type of network has relative advantages, several works have explored combining both architectures into a single network~\cite{salminen2018anatomy}. {Conv-LSTM} is a robust architecture to capture long-term dependencies between features extracted by CNN and found more effective than structures solely based on CNN or LSTM, where the class of a word sequence depends on preceding word sequences. 

However, accurate identification of hate speech in Bengali is still a challenging task. Only a few restrictive approaches~\cite{romim2020hate,ishmam2019hateful,karim2020classification} have been proposed so far. Romim et al.~\cite{romim2020hate} prepared a dataset of 30K comments, making it one of the largest datasets for identifying offensive and hateful statements. However, this dataset has several issues. First, it is very imbalanced as the ratio of hate speech to non-hate speech is 10K:20K. Second, the majority of hate statements are very short in terms of length and word count compared to non-hate statements. Third, their approach exhibits a moderate level of effectiveness at identifying offensive or hateful statements, giving an accuracy of 82\%. Fourth, their approach is a \emph{black-box} method. 
Ismam et al.~\cite{ishmam2019hateful} collected hateful comments from Facebook and annotated 5,126 hateful statements. They classified them into six classes– hate speech, communal attack, inciteful, religious hatred, political comments, and religious comments. Their approach, based on GRU-based DNN, achieved an accuracy of 70.10\%. 

In a recent approach, Karim et al.~\cite{karim2020classification}, provided classification benchmarks for document classification, sentiment analysis, and hate speech detection for the Bengali language. Their approach, by combining fastText embeddings with multichannel Conv-LSTM network architecture, is probably the first work among a few other studies on hate speech detection. Their Conv-LSTM architecture, by combining fastText embeddings, outperformed Word2Vec and GloVe models, since fastText works well with rare words such that even if a word was not seen during the training, it can be broken down into n-grams to get its corresponding embeddings.
All these restrictive approaches are \emph{black-box} methods. On the contrary, interpretable methods put more emphasis on the transparency and traceability of opaque DNN models. With layer-wise relevance propagation~(LRP)~\cite{LRP2}, relevant parts of inputs that caused a result can be highlighted~\cite{holzinger2020measuring}.
To mitigate opaqueness and to improve explainability in hate speech identification, Binny et al.~\cite{mathew2020hatexplain} proposed `HateXplain' - a benchmark dataset for explainable hate speech detection. They observe that high classification accuracy is not everything, but high explainability is also desired. They measure the explainability of an NLP model w.r.t plausibility and faithfulness that are based on human rationales for training~\cite{zaidan2007using}. 

\section{Proposed Approach}
Inspired by SotA approaches and interpretability methods such as sensitivity analysis~(SA)~\cite{saltelli2002sensitivity} and LRP~\cite{LRP2}, we propose {\emph{DeepHateExplainer}} - a novel approach to accurate identification of hate speech in the Bengali. 
Bengali texts are first comprehensively preprocessed, before classifying them into political, personal, geopolitical, and religious hates, by employing an ensemble of different transformer-based neural architectures:  monolingual Bangla BERT-base, multilingual BERT~(mBERT)-cased/uncased, and XLM-RoBERTa. Then, we identify important terms with SA and LRP to provide human-interpretable explanations, covering both global and local explainability. To evaluate the quality of explanations, we measure comprehensiveness and sufficiency. Further, we train several ML~(i.e., LR, NB, KNN, SVM, RF, GBT) and DNN~(i.e., CNN, Bi-LSTM, and Conv-LSTM with word embeddings) baseline models. To the end, \emph{DeepHateExplainer} focuses on algorithmic transparency and explainability, with the following assumptions:

\begin{itemize}
    \item A majority voting-based ensemble from a panel of independent NLP expert or linguists provides fairer and trustworthy prediction than a single expert.
    \item By decomposing the inner logic~(e.g., what terms the model put more attention to) of a black-box model with probing and SA, the opaqueness can be reduced.
    \item By highlighting the most and least important terms, we can generate human-interpretable explanations.
\end{itemize}

\begin{figure*}
    \centering
    \includegraphics[scale=0.72]{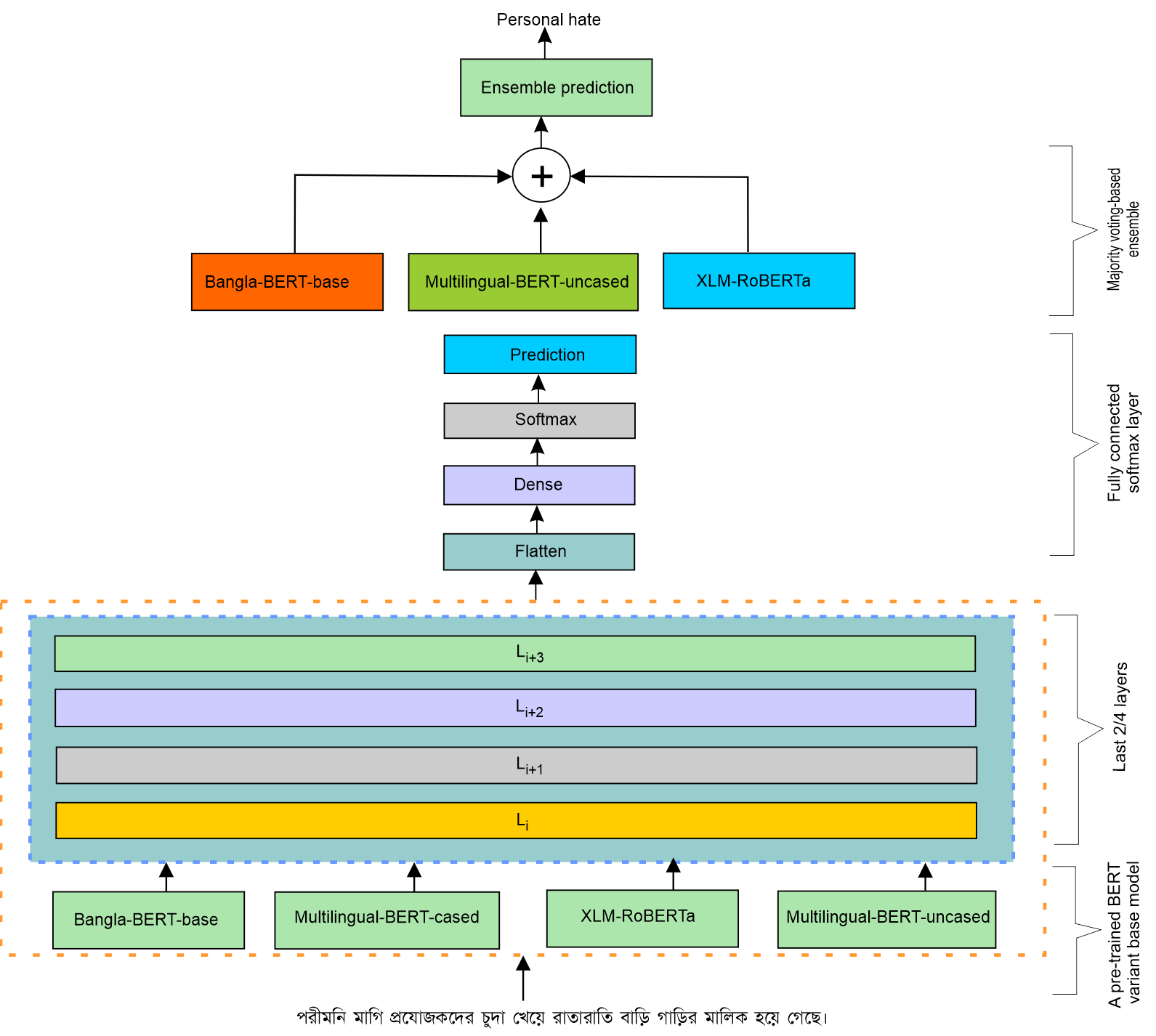}   
    \caption{Schematic representation of proposed approach: each of 4 BERT variants is finetuned by adding a fully-connected softmax layer on top and cross-validation based on ensemble optimization, followed by majority voting ensemble}    
    \label{fig:network_architecture}
\end{figure*}

Overall contributions of our approach are 4-folds:

\begin{enumerate}
    \item We prepared the largest hate speech detection dataset to date for the Bengali language.
    \item To the best of our knowledge, we are the first batch of researchers to employ neural transformer-based language models for hate speech detection for Bengali.
    \item We prepared several computational resources, such as annotated dataset, language models, source codes, and interpretability techniques that will further advance the NLP research for under-resourced Bengali language.
    \item We improved both local and global explainability and algorithmic transparency of \emph{black-box} models by mitigating their opaqueness.
\end{enumerate}

\section{Datasets}\label{section:3}
\begin{figure*}
    \centering
    \includegraphics[scale=0.75]{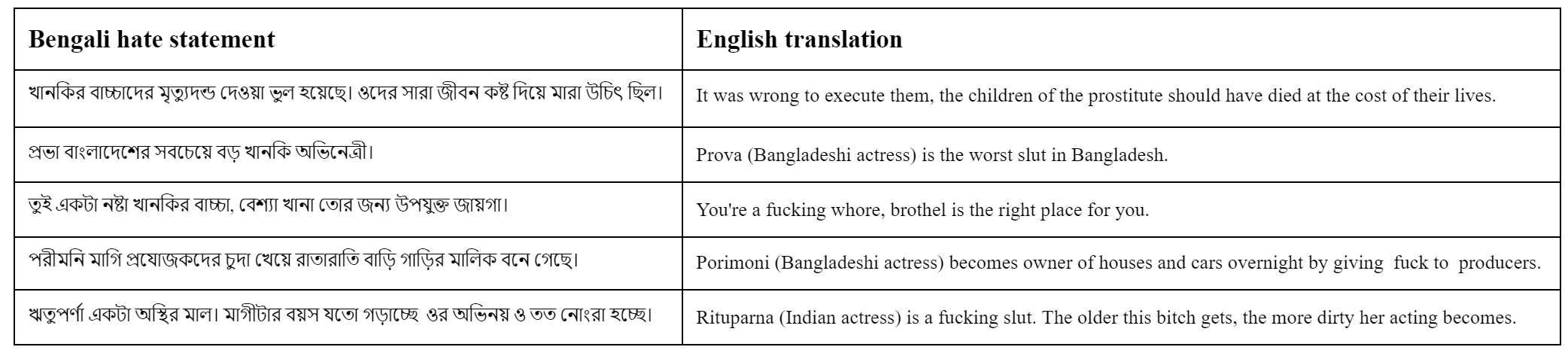}    
    \caption{Example hate statements directed towards a person, but may contextually be directed towards a women}
    \label{fig:personal_vs_gender}
\end{figure*}

We extend the \emph{Bengali Hate Speech Dataset}~\cite{karim2020classification} with additional 5,000 labelled examples. The {Bengali Hate Speech Dataset} categorized observations into political, personal, geopolitical, religious, and gender abusive hates. However, based on our empirical study and linguist analysis, we observe that distinguishing personal from gender abusive hate is often not straightforward, as they often semantically overlap. To justify this, let consider example hate statements in Fig. \ref{fig:personal_vs_gender}. These statements~(non-Bengali speakers are requested to refer to English translations) express hatred statement towards a person, albeit commonly used words such as \includegraphics[width=0.4\textwidth,height=4.5mm]{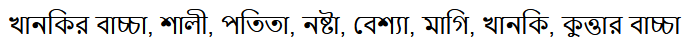}~(corresponding English terms are the girl of slut, slut, prostitute, fucking bitch, whore, waste, bitch), are directed mostly towards women. We follow a bootstrap approach for data collection, where specific types of texts containing common slurs and terms, either directed towards a specific person or entity or generalized towards a group, are only considered. Texts were collected from Facebook, YouTube comments, and newspapers. We categorize the samples into political, personal, geopolitical, and religious hate. Sample distribution and definition of different types of hates are outlined in \Cref{table:stat_hate}.

\subsection{Data annotation}
Three annotators~(a linguist, a native Bengali speaker, and an NLP researcher) participated in the annotation process.
To reduce possible bias, unbiased contents are supplied to the annotators and each label was assigned based on a majority voting on the annotator's independent opinions.
To evaluate the quality of the annotations and to ensure the decision based on the criteria of the objective, we measure inter-annotator agreement w.r.t \emph{Cohen's Kappa} statistic~\cite{chen2005macro}. Let consider $n$ target objects are annotated by $m(\geq 2)$ annotators into one of $k(\geq 2)$ mutually exclusive categories, the proportion of score $\bar{p}_{j}$ and the kappa $\hat{k}_{j}$ for category $j$ are computed as follows~\cite{chen2005macro}:

\begin{align}
    \bar{p}_{j}=\frac{\sum_{i=1}^{n} x_{i j}}{n m}\\
    \hat{k}_{j}=1-\frac{\sum_{i=1}^{n} x_{i j}\left(m-x_{i j}\right)}{n m(m-1) \bar{p}_{j}\left(1-\bar{p}_{j}\right)},
\end{align}

where $x_{ij}$ is possible scores on subject $i$ into category $j$. The overall kappa $\hat{\bar{k}}$ is subsequently computed as~\cite{chen2005macro}:

\begin{equation}
    \hat{\bar{k}}=\frac{\sum_{j=1}^{k} \bar{p}_{j}\left(1-\bar{p}_{j}\right) \hat{k}_{j}}{\sum_{j=1}^{k} \bar{p}_{j}\left(1-\bar{p}_{j}\right)}.
\end{equation}

Taking into account the personal vs. gender abusive hate consideration, we observed a $\hat{\bar{k}}$ score of 0.87, which is 3\% of improvement over the previous approach by Karim et al.~\cite{karim2020classification}.

\begin{table*}
    \centering
    \caption{Statistics of the hate speech detection dataset}
    \label{table:stat_hate}
    \vspace{-4mm}
    \begin{tabular}{l|l|l}
        \hline
        \textbf{Hate type} & \textbf{Description} &  \textbf{\#Examples} \\
        \hline
        Political & Directed towards a political group/party & 999 \\
        \hline
        Religious & Directed towards a religion/religious group & 1,211 \\
        \hline
        Geopolitical & Directed towards a country/region & 2,364 \\
        \hline
        Personal & Directed towards a person & 3,513 \\
        \hline
        \textbf{Total} & &    8,087  \\
        \hline
    \end{tabular}
\end{table*}

\section{Methods}\label{nettwork}
In this section, we discuss our proposed approach in detail, covering word embeddings, network~(ML/DNN/transformers) training, explanation generation, and measuring explainability.

\subsection{Data preprocessing} \label{subsec:preprocessing}
We remove HTML markups, links, image titles, special characters, and excessive use of spaces/tabs, before initiating the annotation process. Further, following preprocessing steps are followed before training ML and DNN baseline models:

\begin{itemize}
    \item \textbf{Hashtags normalization}: inspired by positive effects in classification task~\cite{declerck2015processing}, hashtags were normalized. 
    \item \textbf{Stemming}: inflected words were reduced to their stem, base or root form. 
    \item \textbf{Emojis and duplicates}: all emojis, emoticons, duplicate, and user mentions were removed.
    \item \textbf{Infrequent words}: tokens with a document frequency less than 5 were removed.
\end{itemize}

However, as research has shown that BERT-based models perform better classification accuracy on uncleaned texts, we did not perform major preprocessing tasks, except for the lightweight preprocessing discussed above.

\subsection{Training of ML baseline models}
We train LR, SVM, KNN, NB, RF, and GBT ML baselines models\footnote{Supplementary materials in arXiv version: \url{https://arxiv.org/abs/2012.14353}}, using character n-grams and word uni-grams with TF-IDF weighting. The best hyperparameters are produced through random and with 5-fold cross-validation tests.

\subsection{Neural word embeddings}
We train the \emph{fastText}~\cite{fastText} word embedding model on Bengali articles used for the classification benchmark study by Karim et al.~\cite{karim2020classification}. 
The preprocess reduces vocabulary size due to the colloquial nature of the texts and some degree, addresses the sparsity in the word-based feature representations. We have also tested, by keeping word inflexions, lemmatization, and lower document frequencies. We observe slightly better accuracy using the lemmatization, which is the reason we reported the result based on it.
The fastText model represents each word as an n-gram of characters, which helps capture the meaning of shorter words and allows the embeddings to understand suffixes and prefixes. Each token is embedded into a 300-dimensional real-valued vector, where each element is the weight for the dimension for the token. Since the annotated hate statements are relatively short, we constrain each sequence to 100 words by truncating longer texts and pad shorter ones with zero values to avoid padding in convolutional layers with many blank vectors for the majority of articles.

\subsection{Training of DNN baseline models}
We train three DNN baselines: CNN, Bi-LSTM, and Conv-LSTM. Weights of embedding layer for each network is initialized with the embeddings based on the fastText embedding model. Embedding layer maps each hate statement into a \emph{sequence}~(for LSTM and CNN layers) and transforms into feature representation, which is then flattened and feed into a fully connected softmax layer. 
Further, we add Gaussian noise and dropout layers to improve model generalization. AdaGrad optimizer is used to learn the model parameters by reducing the categorical-cross-entropy loss. We train each DNN architecture 5 separate times in a 5-fold CV setting, followed by measuring the average macro F1-score on the validation set to choose the best hyperparameters\footnote{Supplementary materials in arXiv version: \url{https://arxiv.org/abs/2012.14353}} using random search.

\subsection{Training of transformer-based models}
As shown in Fig. \ref{fig:network_architecture}\footnote{English translation: Porimoni becomes the owner of houses and cars overnight after giving fuck to film producers.}, we train monolingual Bangla BERT-base, mBERT~(cased and uncased), and XLM-RoBERTa large models.  Bangla-BERT-base\footnote{\url{https://huggingface.co/sagorsarker/bangla-bert-base}} is a pretrained Bengali language model built with BERT-based mask language modelling. RoBERTa~\cite{liu2019roberta} is an improved variant of BERT, which is optimized by setting larger batch sizes, introducing dynamic masking, and training on larger datasets. XLM-RoBERTa~\cite{conneau2019unsupervised} is a multilingual model trained on web crawled data. XLM-RoBERTa not only outperformed other transformer models on cross-lingual benchmarks but also performed better on various NLP tasks in a low-resourced language setting.

\begin{table*}[h]
    \centering
    \caption{Hyperparameter combinations for training BERT variants}
    \label{table:bert_params}
    \vspace{-4mm}
    \begin{tabular}{l|l|l|l|l}
        \hline
         \textbf{Hyperparameter}  & \textbf{Bangla-BERT}  &
         \textbf{mBERT cased} &
         \textbf{mBERT-uncased} & \textbf{XLM-RoBERTa}\\
        \hline
        Learning-rate  & 3e-5  & 2e-5 & 5e-5 & 2e-5 \\
        Epochs & 6 & 6 & 6 & 5 \\
        Max seq length & 128 & 128 & 128 & 128 \\
        Dropout & 0.3 & 0.3 & 0.3 & 0.3 \\
        Batch size & 16 & 16 & 16 & 16 \\
        \hline
    \end{tabular}
\end{table*}

\begin{figure*}
    \centering
    \includegraphics[scale=0.57]{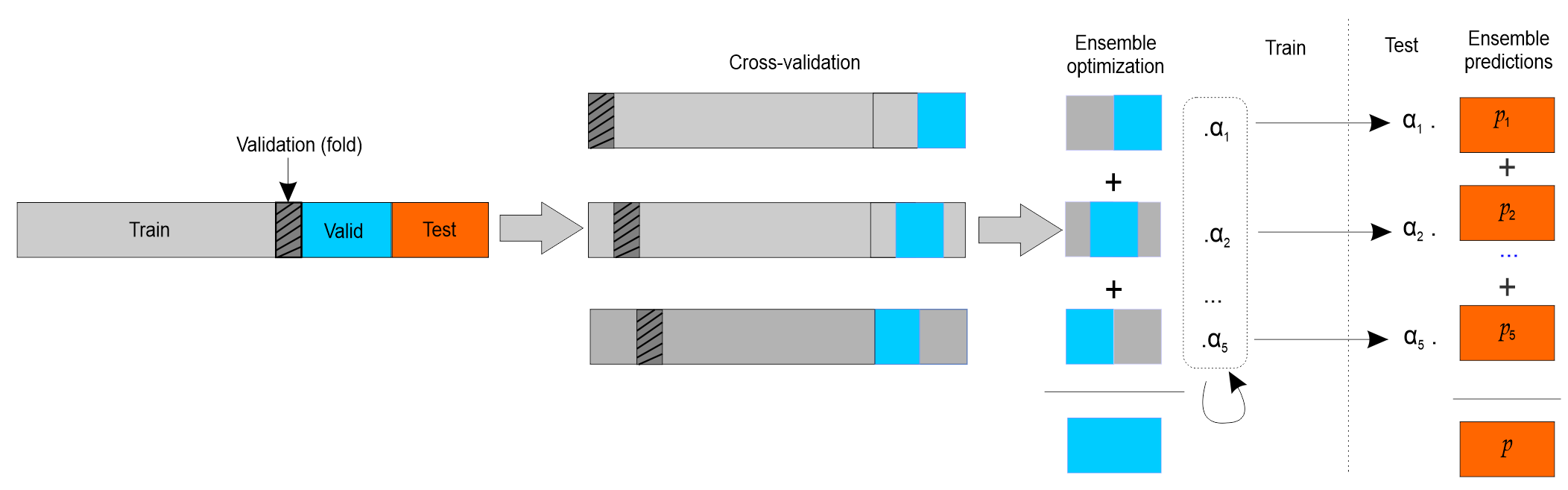}
    \caption{A representation of cross-validation~(CV) ensemble optimization process. The final ensemble weights $\alpha_1, \alpha_2,\cdots, \alpha_M$ in which M is the number of CV folds used to combine model predictions and evaluate performance on test set}
    \label{fig:prediction_ensemble}
\end{figure*}

We shuffle training data for each epoch and apply gradient clipping. We set the initial learning rate to $2 e^{-5}$ and employ Adam optimizer with the scheduled learning rate. Pre-trained BERT variants are fine-tuned by setting the maximum input length to $256$. We experimented with 2, 3, and 4 layers of multi-head attention, followed by a fully connected softmax layer.
As we perform the ensemble of best models to report final predictions~(as Fig. \ref{fig:prediction_ensemble}), several experiments with different hyperparameters combinations are carried out~(\Cref{table:bert_params}), before saving the best epochs, for each model.

\subsection{Generating explanations}
We provide global and local explanations in a post-hoc fashion. For the former, a list of most and least relevant words for each class is identified based on linguist analysis. To provide overall global interpretability, feature importance~(FI) is computed for model $f$.  
For feature $x_i$ in observation $x \in X$ and for each repetition $r$ in $1, 2, \ldots, R$, column $x_i$ is randomly shuffled to generate a corrupted version $\tilde{X}_{r,x_i}$ for $X$. A reference balanced score $s_{r,x_i}$ is then computed for $f$. The mean importance $\sigma_{x_i}$ for feature $x_{i}$ is then computed as follows~\cite{arras2017explaining}:

\vspace{-2mm}
\begin{align}
    \sigma_{x_i}=s-\frac{1}{R} \sum_{r=1}^{R} s_{r,x_i}.
\end{align}

For the latter, we identify which features in a sample are important for individual prediction. Relevance score~(RS) as a measure of importance is computed with SA and relevance conservation LRP~\cite{arras2017explaining}.
For input vector $x$, RS $R_{d}$ is computed for each input dimension $d$. This is analogous to quantify the relevance of $x_{d}$ w.r.t to target class $c$. 
Then the RS $R_{d}$ is generated by computing squared partial derivatives as~\cite{arras2017explaining}:

\begin{align}
    R_{d}=\left(\frac{\partial f_{c}}{\partial x_{d}}(\boldsymbol{x})\right)^{2},
\end{align}

where $f_{c}$ is a prediction score function for class $c$. Total relevances is then computed by summing relevances of all input space dimensions $d$~\cite{arras2017explaining}:

\begin{align}
    \left\|\nabla_{\boldsymbol{x}} f_{c}(\boldsymbol{x})\right\|_{2}^{2}.
\end{align}

In contrast to SA, LRP is based on the layer-wise relevance conservation principle. LRP redistributes the quantity $f_{c}(\boldsymbol{x})$ from output layer to the input layer. The relevance for the output layer neuron is set to $f_{c}(\boldsymbol{x})$ w.r.t to the target class $c$, by ignoring irrelevant output layer neurons. The layer-wise relevance score for each intermediate lower-layer neuron is computed based on weighted connections.
Assuming $z_{j}$ and $z_{i}$ are an upper-layer and a lower-layer neuron, respectively, and the value of $z_{j}$ is already computed in the forward pass as $\sum_{i} z_{i} \cdot w_{ij}+b_{j}$, where $w_{ij}$ and $b_{j}$ are the weight and bias, the relevance score $R_{i}$ for the lower-layer neurons $z_{i}$ is then computed by distributing the relevences onto lower-layer. The relevance propagation $R_{i \leftarrow j}$ from upper-layer neurons $z_{j}$ to lower-layer neurons $z_{i}$ is computed as a fraction of the relevance $R_{j}$. Subsequently, all the incoming relevance for each lower-layer neuron is summed up as~\cite{arras2017explaining}:

\begin{align}
    R_{i \leftarrow j}=\frac{z_{i} \cdot w_{i j}+\frac{\epsilon \cdot \operatorname{sign}\left(z_{j}\right)+\delta \cdot b_{j}}{N}}{z_{j}+\epsilon \cdot \operatorname{sign}\left(z_{j}\right)} \cdot R_{j}
\end{align}

where $N$ is total number of lower-layer neurons connected to $z_{j}$, $\epsilon$ is a stabilizer,  $\operatorname{sign}\left(z_{j}\right)=$ $\left(1_{z_{j} \geq 0}-1_{z_{j}<0}\right)$ is the sign of $z_{j}$, and $\delta$ is a constant multiplicative factor set to 1, to conserve the total relevance of all neurons in the same layer. Finally, $R_{i}$ is computed as $R_{i}= \sum_{j} R_{i \leftarrow j}$~\cite{arras2017explaining}.

\subsection{Measuring explainability}
\emph{System causability scale}~(SCS)~\cite{holzinger2020measuring} is proposed to measure the quality of explanations. SCS is based on the notion of causability and adapted from a usability scale and aims to determine whether and to what extent a user interface is explainable or which explanation process itself is suitable for the intended purpose~\cite{holzinger2020measuring}. Since SCS is based on usability feedback for an explainable interface, it is not suitable for our case. Therefore, we compute faithfulness w.r.t comprehensiveness and sufficiency to measure the quality of explanations based on ERASER~\cite{deyoung2019eraser}.
To measure comprehensiveness, a contrast example $\tilde{x}_{i}$ is created, for each sample $x_{i}$, where $\tilde{x}_{i}$ is calculated by removing predicted rationales $r_{i}$ from $x_{i}$. Let $f\left(x_{i}\right)_{c}$ be the original prediction probability for model $f$ and for predicted class $c$. If model $f$ is defined as $f\left(x_{i} \backslash r_{i}\right)_{c}$ as the predicted probability of $\tilde{x}_{i}\left(=x_{i} \backslash r_{i}\right)$, it is expected that the prediction  will be lower on removing the rationales~\cite{deyoung2019eraser}. The comprehensiveness metric $e$ is then calculated as follows~\cite{deyoung2019eraser}:

\begin{align}
    e  = f\left(x_{i}\right)_{c}-f\left(x_{i} \backslash r_{i}\right)_{c}
\end{align}

\hspace*{3.5mm} The concept of rationales is proposed by Zaidan et al.~\cite{zaidan2007using} in NLP in which human annotators would highlight a span of text that could support their labelling decision, e.g., to justify why a review is positive, an annotator can highlight most important words and phrases that would tell someone to see the movie. To justify why a review is negative, highlight words and phrases that would tell someone not to see the movie. It is found to be useful in downstream NLP tasks like hate speech detection~\cite{mathew2020hatexplain}, text classification~\cite{herrewijnenmachine}. We conceptualize a similar idea w.r.t \emph{leave-one-feature-out} analysis, where the rationale is computed based on the number of highlighted features divided by the number of features in a test sample. A prediction is considered a match if it overlaps with any of the ground truth rationales $r_i \geq 0.5$. A high value of comprehensiveness implies that the rationales were influential in the prediction. The sufficiency $s$, which measures the degree to which extracted rationales are adequate for the model $f$, which is measured as follows~\cite{deyoung2019eraser}:

\begin{align}
    s  =  f\left(x_{i}\right)_{c} f\left(r_{i}\right)_{c}
\end{align}

\section{Results}\label{er}
We discuss experimental results both qualitatively and qualitatively and explain the predictions globally and locally. Besides, we provide a comparative analysis with baselines.

\subsection{Experiment setup}
Programs were implemented using \emph{scikit-learn}, \emph{Keras}, and \emph{PyTorch} and networks are trained on Nvidia GTX 1050 GPU. 
Open source implementation of \emph{fastText}\footnote{\url{https://radimrehurek.com/gensim/models/fasttext.html}} is used to learn embeddings. \emph{SHAP}\footnote{\url{https://github.com/slundberg/shap}} and \emph{ELI5}\footnote{\url{https://github.com/eli5-org/eli5}} are used to compute FI. Each model is trained on 80\% of data, followed by evaluating the
model on 20\% held-out data. We report precision, recall, F1-score, and \emph{Matthias correlation coefficient}~({MCC}). 
Finally, we perform the ensemble of top-3 models to report the final predictions. We select the best models with \emph{WeightWatcher}\footnote{\url{https://github.com/CalculatedContent/WeightWatcher}}~\cite{martin2019traditional}. Using WeightWatcher, the models giving the lowest \emph{log-norm} and highest \emph{weighted-alpha} are only considered. This is backed by the fact that a lower log-norm signifies better generalization of network weights for unseen examples~\cite{martin2019traditional}.

\begin{table*}
            \centering
            \caption{Performance of hate speech detection}
            \label{table:hate_result_v2}
            \vspace{-4mm}
            \begin{tabular}{c|l|l|l|l|l}
            \hline
            \textbf{Method} & \textbf{Classifier} & \textbf{Precision} & \textbf{Recall} & \textbf{F1} & \textbf{MCC}\\ \hline
             \multirow{6}{*}{ML baselines}
            & LR & 0.68 &    0.68 &    0.67 & 0.542 \\
             & NB & 0.65 &    0.65 &    0.64 &    0.511 \\
             & SVM & 0.67 &    0.67 &    0.66 &    0.533\\
             & KNN & 0.67 &    0.67 &    0.66 &    0.533\\
             & RF & 0.69 &    0.69 &    0.68 &    0.561\\
             & GBT & \textbf{0.71} &    \textbf{0.69} &    \textbf{0.68} &    \textbf{0.571} \\
             \hline
             \multirow{3}{*}{DNN baselines}
             & CNN & 0.74 &    0.73 &    0.73 &    0.651\\
             & Bi-LSTM & 0.75 &    0.75 &    0.75 &    0.672\\
             & Conv-LSTM & \textbf{0.79} &\textbf{0.78} &    \textbf{0.78} & \textbf{0.694}\\
             \hline
            \multirow{5}{*}{BERT variants} & Bangla BERT & 0.86  &    0.86  &    0.86 &    0.799\\
             & mBERT-cased & 0.85  &    0.85   &   0.85 &    0.774 \\
             & XML-RoBERTa & \textbf{0.87}  &   \textbf{0.87 } & \textbf{0.87}  &   \textbf{0.808} \\
             & mBERT-uncased & 0.86    &  0.86   &   0.86  &    0.795\\
             & Ensemble* & \textbf{0.88} & \textbf{0.88}  &    \textbf{0.88} &    \textbf{0.820} \\ \hline
            \end{tabular}
\end{table*}

\begin{table}
            \centering
            \caption{Class-wise classification report based on\\ majority voting ensemble of top-3 classifiers}
            \label{table:class_specific_result_v2}
            \vspace{-4mm}
            \begin{tabular}{l|l|l|l}
            \hline
            \textbf{Hate type}  &
            \textbf{Precision}  &
            \textbf{Recall} &
            \textbf{F1} \\
            \hline
            Personal & 0.91 & 0.90 & 0.91\\
            Political & 0.82 & 0.74 & 0.78\\
            Religious & 0.79 & 0.90 & 0.84\\
            Geopolitical & 0.89 & 0.89 & 0.89\\
            \hline
            \end{tabular}
\end{table}

\subsection{{Analysis of hate speech detection}}
\label{seq:result_analysis}
We evaluated 4 variants of BERT models on the held-out test set and report the results\footnote{Based on hyperparameter combinations in \Cref{table:bert_params}.} in \Cref{table:hate_result_v2}. XML-RoBERTa model turns out to be both best performing and best-fitted model, giving the top F1-score of 87\%, which is about 2\% to 5\% better than other transformer models, while Bangla BERT-base and mBERT-uncased also performed moderately well. Based on metrics and the lowest log-norm, top-3 models were picked using WeightWatcher for the ensemble prediction, followed by discarding the mBERT-cased model from the voting ensemble. The highest MCC score of 0.82 is achieved with the ensemble prediction, which is slightly better than that of the XLM-RoBERTa, giving an MCC score of 0.808. Overall, MCC scores of $\geq 0.77$ were observed for each BERT-based model w.r.t Pearson correlation coefficient. This signifies that predictions are strongly correlated with ground truths and BERT variants are more effective compared to ML or DNN baseline models.

\begin{figure*}
    \centering
    \begin{subfigure}{.49\linewidth}
        \centering
        \includegraphics[scale=0.57]{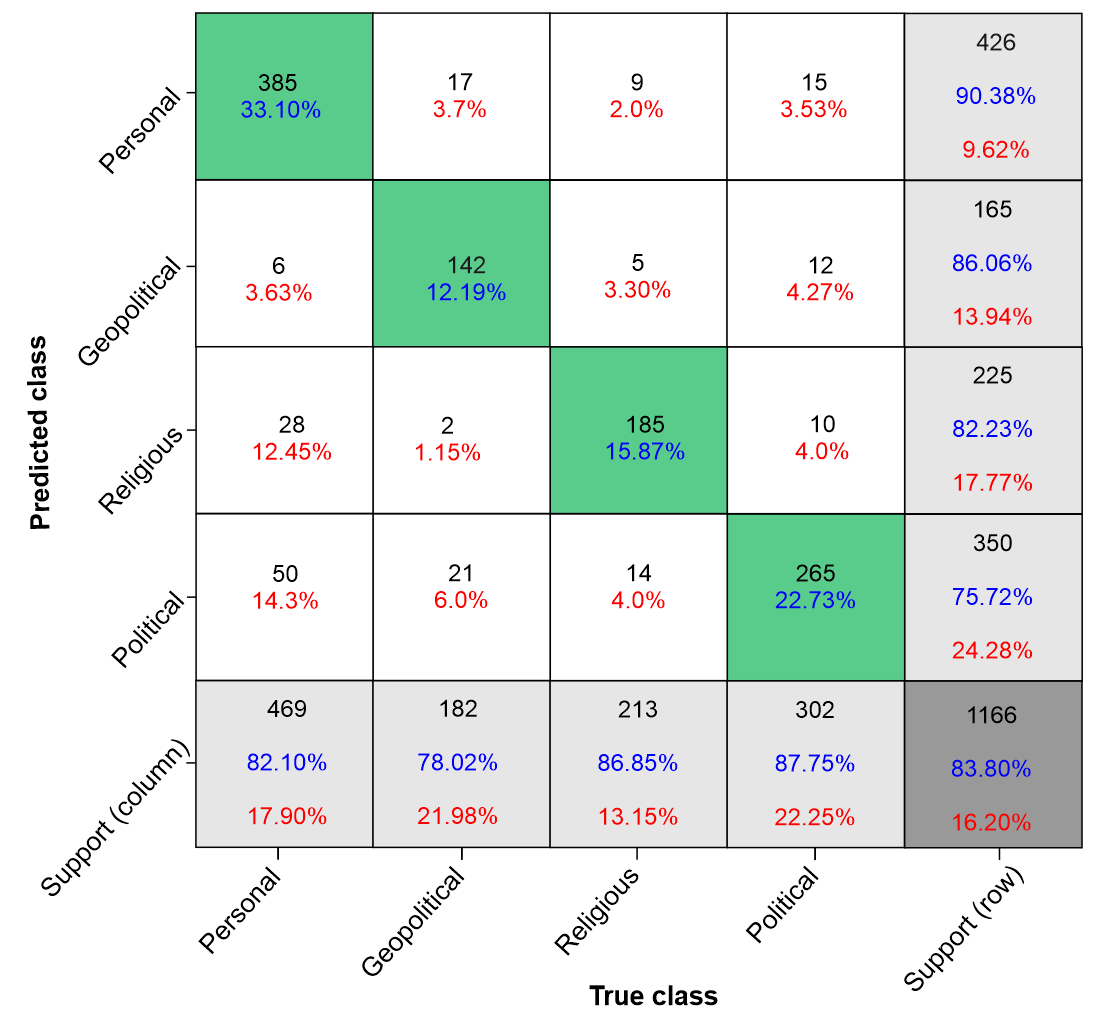}
        \caption{For standalone XLM-RoBERTa}
        \label{fig:conf_5_classes}
    \end{subfigure}
    \begin{subfigure}{.49\linewidth}
        \centering
        \includegraphics[scale=0.57]{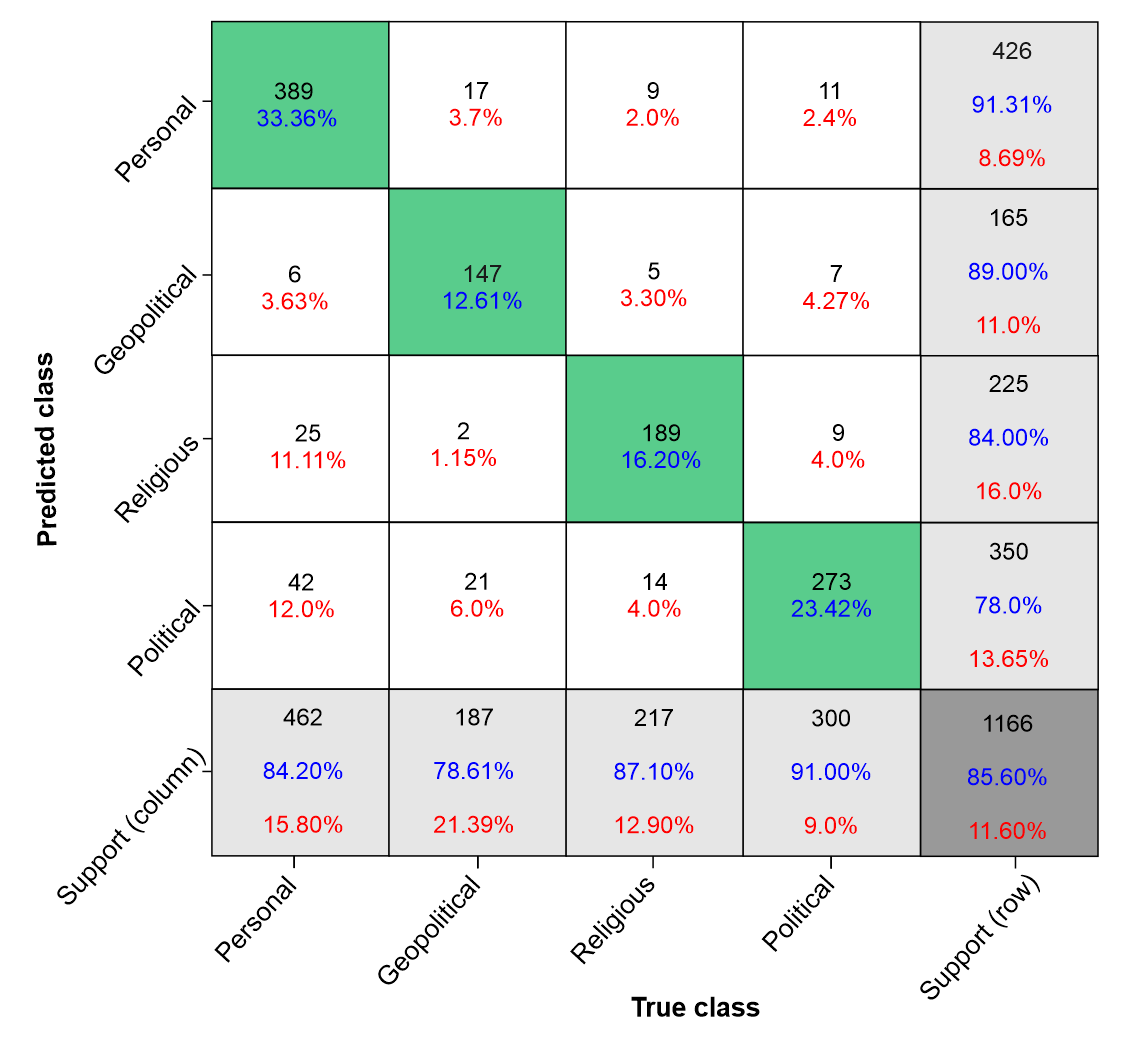}
        \caption{For ensemble prediction}
        \label{fig:conf_4_classes}
    \end{subfigure}
    \caption{Confusion matrices: standalone XLM-RoBERTa vs. ensemble prediction~(color code: red, blue, and black indicate\\ misclassification rates, correct classification rates~(in \%), and count, respectively)}
    \label{fig:conf_matrices}
\end{figure*}

Confusion matrices in Fig. \ref{fig:conf_matrices} show the breakdown of correct and incorrect classifications for each class, which correspond to ground truths vs. predicted labels. 
Ensemble prediction boosts the accuracy by at least 1.8\% across the classes w.r.t F1-score, compared to top mBERT-cased and XML-RoBERTa models. Nevertheless, misclassification rates for all the classes have reduced significantly and overall 21 observations were correctly classified. This improvement signifies, to large extent, that ensemble prediction is effective at minimizing confusions.
Further, as classes are imbalanced, accuracy alone gives a distorted estimation of the performance. Thus, we provide class-specific classification reports in \Cref{table:class_specific_result_v2} based on the ensemble prediction. Overall, our approach identifies personal hates more accurately compared to other types of hate w.r.t F1-score. Identifying political hate was more challenging~(giving an F1-score of 0.78) as political hates contain some terms that are often used to express personal hates.

\begin{figure*}
    \centering
    \frame{\includegraphics[scale=0.87]{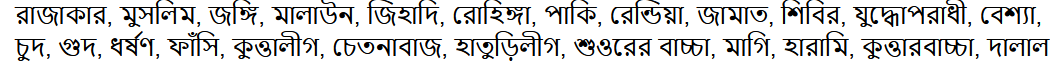} }   
    \caption{Globally most important terms that are used to express hatred statements for all the hate classes}    
    \label{fig:global_most_term}
\end{figure*}

\subsection{Comparison with baselines}\label{seq:result_baselines}
Since efficient feature selection can have significant impacts on model performance for ML methods~\cite{karim2020classification}, we observe the performance with manual feature selection. Forests of trees concept\footnote{Forests of trees concept is a meta-transformer for selecting features w.r.t importance weights.} is employed to compute impurity-based FI. Each model is then trained by discarding irrelevant features. 
The feature selection helped SVM, KNN, RF, and GBT models improve their accuracy. 
GBT model performs the best among all ML baseline models, giving an MCC score of 0.571, albeit F1-scores for both RF and GBT are equal. RF model performs reasonably well, giving an F1-score of 68\%. Contrarily, performance of SVM, LR, and NB classifiers degraded significantly. LR model is not resilient to class discriminating features that could be lost during the feature selection, perhaps the conditional independence assumption~(where features are assumed to be independent when conditioned upon class labels) of NB is not hold. Overall, the performance of each ML baseline model was severely poor, making them not suitable for reliable identification of hate statements.

Each DNN baseline model is evaluated by initializing the embedding layer's weight with {fastText} embeddings. As observed, each model either outperforms or gives comparable performance to ML baseline models. In particular, Conv-LSTM performs the best among DNN 
baselines, giving F1 and MCC scores of 0.78 and 0.694, respectively, which is about 4 to 5\% better than Bi-LSTM~(the second-best DNN baseline) and GBT~(the best among ML baseline) models, respectively; while the F1-scores for CNN and Bi-LSTM reached to 0.73 and 0.75, respectively, making them comparable to GBT and RF models. Overall, DNN baseline models also performed poorly compared to transformer-based models~(ref. \Cref{table:hate_result_v2}), albeit the fastText embedding model could have captured the word-level semantics sufficiently.

\begin{figure*}
    \centering
    \includegraphics[scale=0.73]{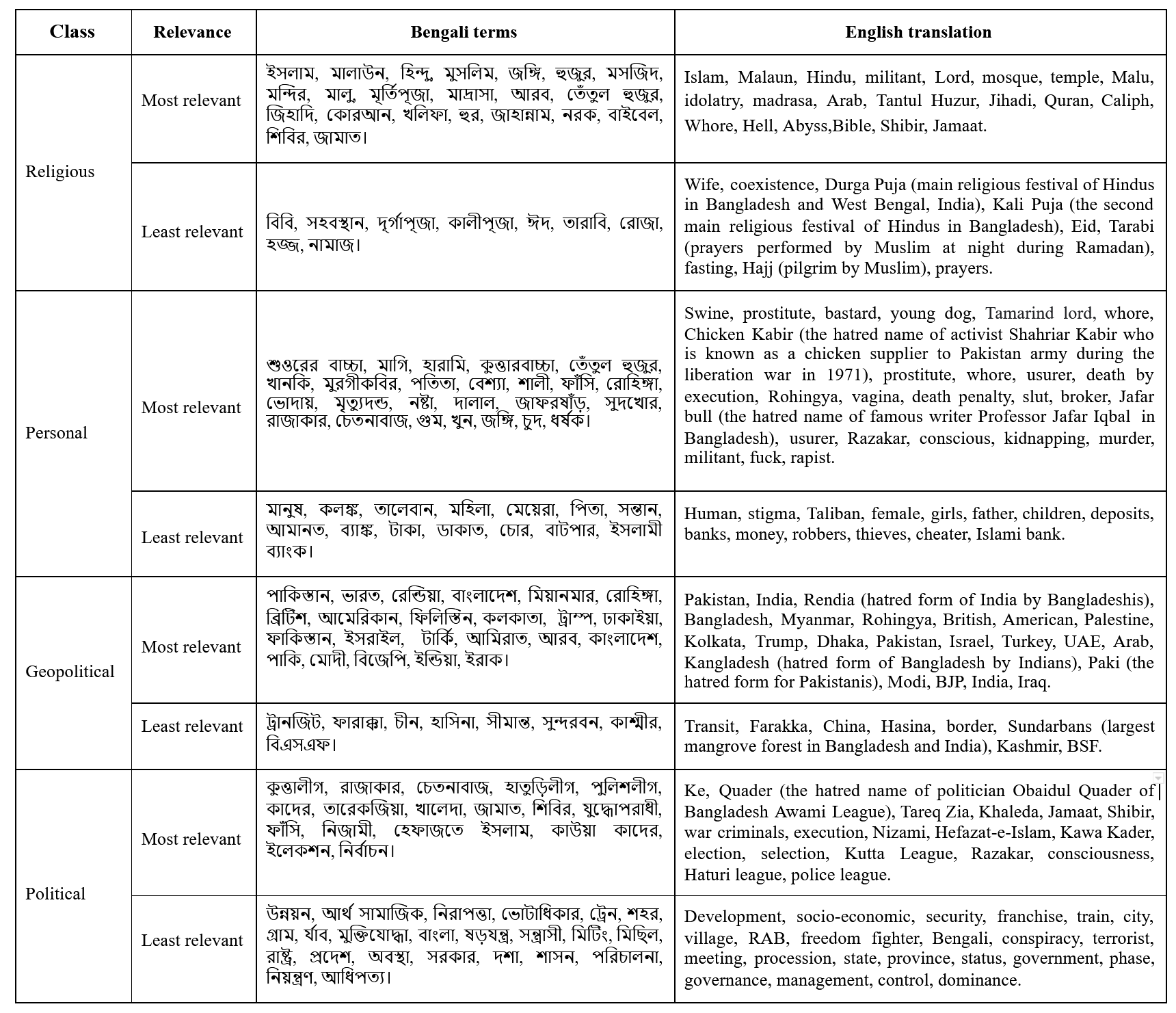}    
    \caption{Globally most important terms used to express hatred statements for each hate\\ class and their relevance interpretation}    
    \label{fig:global_most_term_per_class}
\end{figure*}
\begin{figure*}
    \centering
    \includegraphics[scale=0.65]{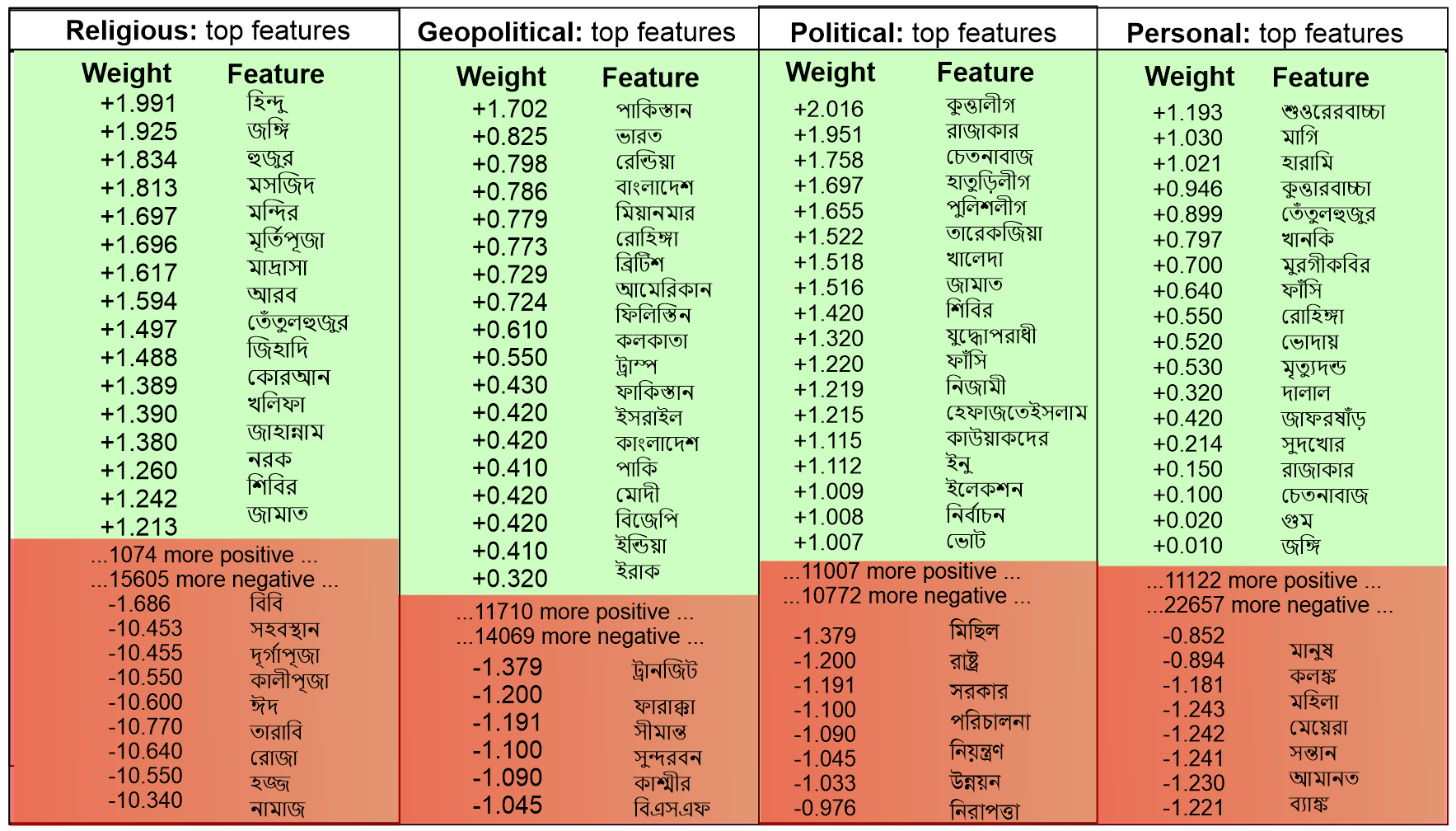}    
    \caption{Global feature importance, highlighting important terms per class}
    \label{fig:global_hateful_words_per_class}
\end{figure*}

\subsection{Explaining hate speech detection}
We provide both local and global explanations for hate speech identification. For the former, we highlight globally important terms. 
Fig. \ref{fig:global_most_term} shows most frequently used terms expressing hatred statements~(English terms: Rajakar, war criminals, Muslim, militant, Hindu, Jihadi\footnote{Term to accuse Muslims to be terrorist in India, Pakistan, and Bangladesh.}, Rohinga\footnote{People who flew from genocide and ethnic cleansing by the Myanmar army and got asylum in Bangladesh.}, Pakistanis, Indians, Bangladesh Jamaat-e-Islami\footnote{Islamist political party in Bangladesh.}, war criminals, whore, fuck, ass, rape, execution, Kutta League\footnote{Hatred term for student organization of Bangladesh Awami League, where Kutta means dogs.}, consciousness\footnote{The hatred form for Bangladesh Awami League, whose political agenda is backed by liberation war.}, Hammer League\footnote{The hatred term of the student league - the official student organization of Bangladesh Awami League, who are suspects of killing many oppositions and innocent people with a hammer and hock-stick.}, son of a pig, slut, bastard, son of a bitch, broker\footnote{Supporters of Bangladesh Awami League are called brokers of India, while supporters of Bangladesh Nationalist Party and Bangladesh Jamaat-e-Islami are called brokers of Pakistan.}). 
These findings are further validated with the linguistic analysis, outlining the semantic meaning and relevance of these words. The most and least SA- and LRP-relevant word lists for each class are shown in Fig. \ref{fig:global_hateful_words_per_class} and Fig. \ref{fig:global_most_term_per_class}, respectively that are used to express hatred statements.

\begin{figure*}
    \centering
    \begin{subfigure}{.49\linewidth}
        \centering
        \includegraphics[scale=0.52]{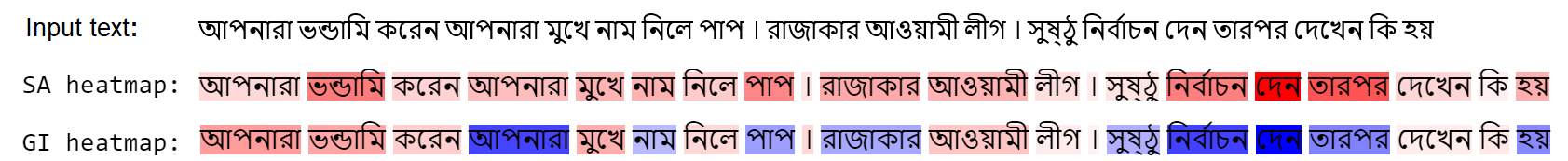}
        \caption{Political hate}
        \label{fig:poli_hate_explain}
    \end{subfigure}\hfill
    \begin{subfigure}{0.49\linewidth}
        \centering
        \includegraphics[scale=0.5]{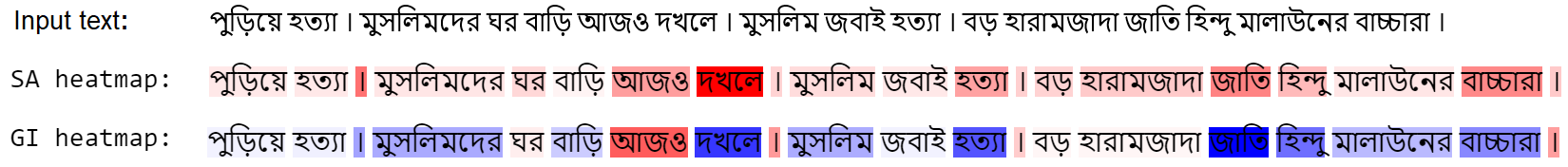}
        \caption{Religious hate}
        \label{fig:reli_hate_explain}
    \end{subfigure}\hfill
    \begin{subfigure}{.49\linewidth}
        \centering
        \includegraphics[scale=0.55]{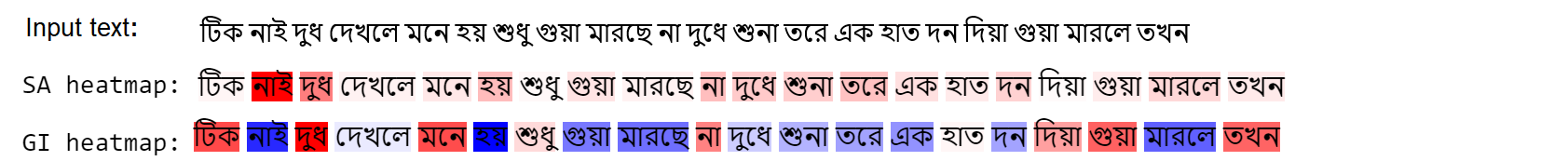}
        \caption{Personal hate}
        \label{fig:per_hate_explain}
    \end{subfigure}\hfill
    \begin{subfigure}{0.49\linewidth}
        \centering
        \includegraphics[scale=0.55]{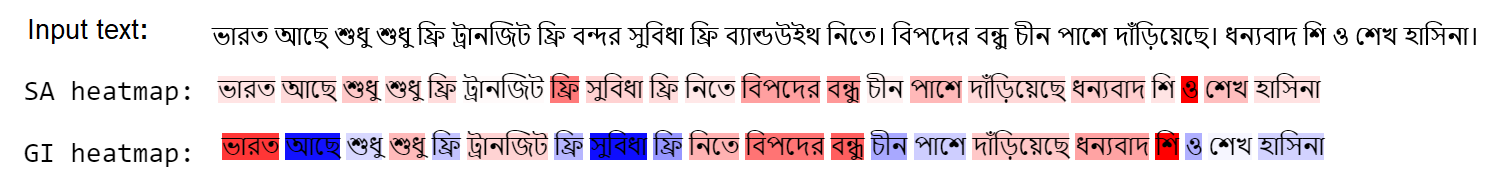}
        \caption{Geopolitical hate}
        \label{fig:geo_hate_explain}
    \end{subfigure}\hfill
    \caption{Example heat maps for for different types of hate, highlighting relevant terms}
    \label{fig:class_wise_local}
\end{figure*}

Local explanations for individual samples are provided by highlighting the most important terms. We provide class-wise example heat maps based on SA and LRP-based relevances in Fig. \ref{fig:class_wise_local} exposing different types of hates, where the colour intensity is normalized to the maximum relevance per hate statement. To quantitatively validate the word-level relevances for local explainability, we perform the leave-one-out experiment -- we aim to improve the greedy backward elimination algorithm by preserving more interactions among terms. First, we randomly select a sample hate statement~(e.g., same as Fig. \ref{fig:reli_hate_explain}) in the test set. Then, we generate prediction probabilities for all the classes, followed by explaining word-level relevance for the two highest probable classes.

\begin{figure}
    \centering
    \includegraphics[scale=0.5]{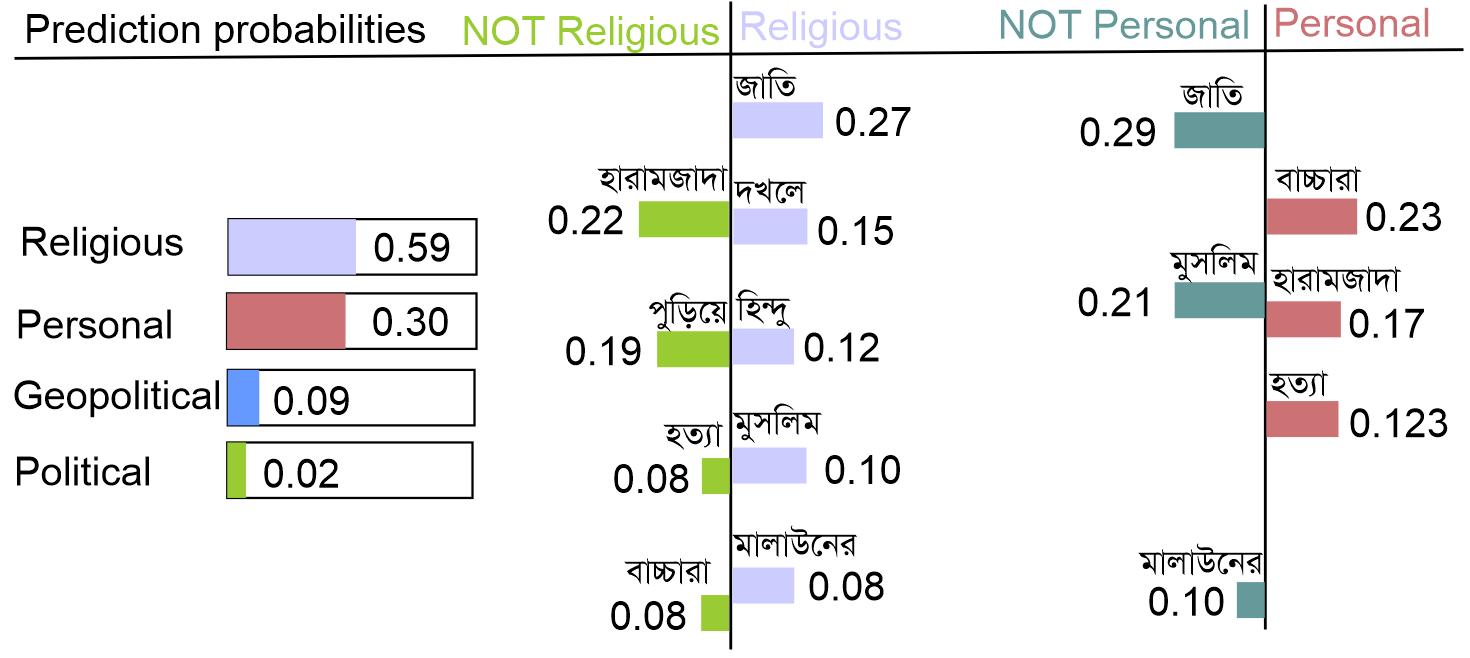}
    \caption{Word-level relevance test using leave-one-out}
    \label{fig:explain_term_wise_religious_hate}
\end{figure}

Let consider the example in Fig. \ref{fig:explain_term_wise_religious_hate}: words on the right side are positive, while words on the left are negative. Words like  \includegraphics[width=0.15\textwidth,height=0.9\baselineskip]{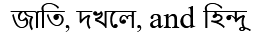}~(race, Occupy, and Hindu in English, respectively) are positive for religious class, albeit the most significant word \includegraphics[width=0.045\textwidth,height=4mm]{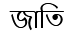}~(race in English) is negative for personal hate category~(where words \includegraphics[width=0.15\textwidth,height=0.9\baselineskip]{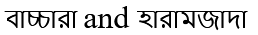}~(son of a bitch and bastard in English) are more important). Word \includegraphics[width=0.045\textwidth,height=4mm]{inline_5.png} has the highest positive score of 0.27 for class religious. Our model predicts this as a religious hate statement too, with the probability of 59\%.
However, if we remove word \includegraphics[width=0.045\textwidth,height=4mm]{inline_5.png} from the text, we would expect the model to predict the label religious with a probability of 32\%~(i.e., 59\% $-$ 27\%). Word \includegraphics[width=0.045\textwidth,height=4mm]{inline_5.png} is negative for personal hate category, albeit words \includegraphics[width=0.15\textwidth,height=0.9\baselineskip]{inline_4.png} have positive scores of 0.23 and 0.17 for the class personal. These identified words not only reveals the relevance of important terms for classifier's decision, but also signify that removing most relevant terms will impact the final decision, accordingly to their relevance value.

\subsection{Measure of explainability}
For measuring the explainability, only top models~(ML, DNN, and BERT variants) are considered based on the results we analyzed in \Cref{seq:result_analysis} and \Cref{seq:result_baselines}. Results of the faithfulness in terms of comprehensiveness and sufficiency are shown in \Cref{tab:faithfullness_result}. As shown, XML-RoBERTa attained the highest comprehensiveness and sufficiency scores, outperforming other standalone models. Overall, BERT variants not only attained higher scores but also consistently outperforms other models such as GBT and Conv-LSTM baselines. Further, our study outlines two additional observations:

\begin{enumerate}
    \item GBT model shows both higher comprehensiveness and sufficiency compared to Conv-LSTM model, albeit the latter outperformed the former in classification task w.r.t classification metrics.
    \item As for BERT variants, Bangla BERT and mBERT-cased generate the least faithful explanations.
\end{enumerate}

\begin{table}
    \centering
    \caption{Measure of explainability}
    \vspace{-4mm}
    \begin{tabular}{l|c|c}
        \hline \textbf{Classifier} & \textbf{Comprehensiveness} & \textbf{Sufficiency} \\
        \hline GBT & 0.79 & 0.25 \\ \hline
        Conv-LSTM & 0.73 & 0.15 \\
        \hline Bangla BERT & 0.78 & 0.25 \\  
        \hline XML-RoBERTa & 0.84 & 0.44 \\ \hline
        mBERT-uncased & 0.81 & 0.35 \\
        mBERT-cased & 0.76 & 0.28 \\
        \hline
    \end{tabular}
    \label{tab:faithfullness_result}
\end{table}

This signifies that a model that attains the best scores w.r.t metrics, may not perform well in terms of faithfulness explainability metrics. Based on this observation, it would not be unfair to say that a model's performance metric alone is not enough as models with slightly lower performance, but much higher scores for faithfulness might be preferred for sensitive use cases such as hate speech detection at hand.

\section{Conclusion}\label{con}
In this paper, we proposed \emph{DeepHateExplainer} - an explainable approach for hate speech detection for under-resourced Bengali language. Based on ensemble prediction, \emph{DeepHateExplainer} can detect different types of hates with an F1-score of 88\%, outperforming several ML and DNN baselines. Our study suggests that: i) feature selection can have non-trivial impacts on learning capabilities of ML and DNN models, ii) even if a standalone ML and DNN baseline model does not perform well, the ensemble of several models may still outperform individual models. 

Our approach has several potential limitations too. First, we had a limited amount of labelled data at hand during the training. Therefore, it would be unfair to claim that we could rule out the chance of overfitting. Secondly, we applied SA and LRP on a DNN baseline model~(i.e., Conv-LSTM), albeit it would be more reasonable to do the same on the best performing standalone XLM-RoBERTa model. In future, we want to overcome these limitations by extending the datasets with a substantial amount of samples and applying SA and LRP on the XLM-RoBERTa model. Besides, we want to focus on other interesting areas such as named entity recognition, part-of-speech tagging, sense disambiguation, and question answering for the Bengali language.    

\printbibliography

@article{1,
	title={Diagnosis of OA: imaging},
	author={Braun, Hillary J and Gold, Garry E},
	journal={Bone},
	volume={51},
	number={2},
	pages={278--288},
	year={2012},
	publisher={Elsevier}
}

@inproceedings{2,
	title={Severity analysis of Osteoarthritis of knee joint from X-ray images: A Literature review},
	author={Kawathekar, Pooja P and Karande, Kailash J},
	booktitle={Signal Propagation and Computer Technology (ICSPCT), 2014 International Conference on},
	pages={648--652},
	year={2014},
	organization={IEEE}
}

@article{3,
	title={Hip and knee arthroplasty in the geriatric population},
	author={Clair, Selvon F St and Higuera, Carlos and Krebs, Viktor and Tadross, Nabil A and Dumpe, Jerrod and Barsoum, Wael K},
	journal={Clinics in geriatric medicine},
	volume={22},
	number={3},
	pages={515--533},
	year={2006},
	publisher={Elsevier}
}

@article{5,
	title={The effects of specific medical conditions on the functional limitations of elders in the Framingham Study.},
	author={Guccione, Andrew A and Felson, David T and Kannel, William B},
	journal={American journal of public health},
	volume={84},
	number={3},
	pages={351--358},
	year={1994},
	publisher={American Public Health Association}
}

@inproceedings{6,
	title={Socio-economic costs of osteoarthritis: a systematic review of cost-of-illness studies},
	author={Puig-Junoy, Jaume and Zamora, Alba Ruiz},
	booktitle={Seminars in arthritis and rheumatism},
	volume={44},
	number={5},
	pages={531--541},
	year={2015},
	organization={Elsevier}
}

@article{LRP2,
  title={Explaining Convolutional Neural Networks using Softmax Gradient Layer-wise Relevance Propagation},
  author={Iwana, Brian Kenji and Kuroki, Ryohei and Uchida, Seiichi},
  journal={arXiv:1908.04351},
  year={2019}
}

@article{martin2019traditional,
  title={Traditional and heavy-tailed self regularization in neural network models},
  author={Martin, Charles H and Mahoney, Michael W},
  journal={arXiv:1901.08276},
  year={2019}
}

@article{conneau2019unsupervised,
  title={Unsupervised cross-lingual representation learning at scale},
  author={Conneau, Alexis and Khandelwal, Kartikay and Goyal, Naman and Chaudhary, Vishrav and Wenzek, Guillaume and Guzm{\'a}n, Francisco and Grave, Edouard and Ott, Myle and Zettlemoyer, Luke and Stoyanov, Veselin},
  journal={ arXiv:1911.02116},
  year={2019}
}

@article{liu2019roberta,
  title={{RoBERTa}: A robustly optimized {BERT} pretraining approach},
  author={Liu, Yinhan and Ott, Myle and Goyal, Naman and Lewis, Mike and Zettlemoyer, Luke and Stoyanov, Veselin},
  journal={ arXiv:1907.11692},
  year={2019}
}

@inproceedings{yang2019xlnet,
  title={{XLNET}: Generalized autoregressive pretraining for language understanding},
  author={Yang, Zhilin and Dai, Zihang and Yang, Yiming and Carbonell, Jaime and Salakhutdinov, Russ R and Le, Quoc V},
  booktitle={Advances in neural information processing systems},
  pages={5753--5763},
  year={2019}
}

@article{clark2020electra,
  title={Electra: Pre-training text encoders as discriminators rather than generators},
  author={Clark, Kevin and Luong, Minh-Thang and Le, Quoc V and Manning, Christopher D},
  journal={ arXiv:2003.10555},
  year={2020}
}

@book{karim,
    title = {Predictive Analytics with TensorFlow},
    author = {Md Rezaul Karim},
    isbn = {9781788398923},
    year = {2017},
    publisher = {Packt Publishing Ltd.}
}

@inproceedings{declerck2015processing,
  title={Processing and normalizing hashtags},
  author={Declerck, Thierry and Lendvai, Piroska},
  booktitle={Proceedings of the International Conference Recent Advances in Natural Language Processing},
  pages={104--109},
  year={2015}
}

@inproceedings{SHAP,
  title={A unified approach to interpreting model predictions},
  author={Lundberg, Scott M and Lee, Su-In},
  booktitle={Advances in neural information processing systems},
  pages={4765--4774},
  year={2017}
}

@inproceedings{ishmam2019hateful,
  title={Hateful Speech Detection in Public Facebook Pages for the {Bengali} Language},
  author={Ishmam, Alvi Md and Sharmin, Sadia},
  booktitle={2019 18th IEEE International Conference On Machine Learning And Applications~(ICMLA)},
  pages={555--560},
  year={2019},
  organization={IEEE}
}

@article{arras2017explaining,
  title={Explaining recurrent neural network predictions in sentiment analysis},
  author={Arras, Leila and Montavon, Gr{\'e}goire and Muller, Klaus-Robert and Samek, Wojciech},
  journal={ arXiv:1706.07206},
  year={2017}
}

@inproceedings{elsherief2018hate,
  title={Hate lingo: A target-based linguistic analysis of hate speech in social media},
  author={Sherief, Mai and Kulkarni, Vivek and Belding, Elizabeth},
  booktitle={12th AAAI Conference on Web and Social Media},
  year={2018}
}

@inproceedings{ribeiro2018characterizing,
  title={Characterizing and detecting hateful users on {Twitter}},
  author={Ribeiro, Manoel Horta and Calais, Pedro H and Almeida, Virg{\'\i}lio AF and Meira Jr, Wagner},
  booktitle={12th AAAI conference on web and social media},
  year={2018}
}

@article{hate5,
  title={Hate speech and incitement to hatred against minorities in the media},
  author={Izs´ak, R.},
  journal={UN Humans Rights Council, A/HRC/28/64},
  year={2015}
}

@inproceedings{zhang2018detecting,
  title={Detecting Hate Speech on {Twitter} Using a Convolution-{GRU} Based Neural Network},
  author={Zhang, Ziqi and Robinson, David and Tepper, Jonathan},
  booktitle={ESWC},
  pages={745--760},
  year={2018},
  organization={Springer}
}

@inproceedings{salminen2018anatomy,
  title={Anatomy of Online Hate: Developing a Taxonomy and ML Models for Identifying and Classifying Hate in Online News Media.},
  author={Salminen, Joni and Almerekhi, Hind and Milenkovic, Milica and Jung},
  booktitle={ICWSM},
  pages={330--339},
  year={2018}
}

@article{devlin2018BERT,
  title={{BERT}: Pre-training of deep bidirectional transformers for language understanding},
  author={Devlin, Jacob and Chang, Ming-Wei and Lee, Kenton and Toutanova, Kristina},
  journal={ arXiv:1810.04805},
  year={2018}
}

@inproceedings{chen2005macro,
  title={A macro to calculate kappa statistics for categorizations by multiple raters},
  author={Chen, Bin and Zaebst, Dennis and Seel, Lynn},
  booktitle={Proceeding of the 30th Annual SAS Users Group International Conference},
  pages={155--30},
  year={2005},
  organization={Citeseer}
}

@misc{Sagor_2020,
  title   = {{Bangla-BERT}: {Bengali} Mask Language Model for {Bengali} Language Understading},
  author  = {Sagor Sarker},
  year    = {2020},
  url    = {https://github.com/sagorbrur/bangla-bert}
}

@inproceedings{fastText,
  title={Learning Word Vectors for 157 Languages},
  author={Grave, Edouard and Bojanowski, Piotr and Mikolov, Tomas},
  booktitle={Proc. of the International Conference on Language Resources and Evaluation~(LREC)},
  year={2018}
}

@inproceedings{islam2009research,
  title={Research on {Bangla} language processing in {Bangladesh}: progress and challenges},
  author={Islam, MS},
  booktitle={8th International Language and Development Conference},
  pages={23--25},
  year={2009}
}

@inproceedings{karim2020classification,
  title={Classification benchmarks for under-resourced {Bengali} language based on multichannel convolutional-{LSTM} network},
  author={Karim, Md Rezaul and Chakravarthi, Bharathi Raja and McCrae, John P and Cochez, Michael},
  booktitle={2020 IEEE 7th International Conference on Data Science and Advanced Analytics (DSAA)},
  pages={390--399},
  year={2020},
  organization={IEEE}
}

@article{guterres2019united,
  title={United Nations Strategy and Plan of Action on Hate Speech},
  author={Guterres, A},
  journal={Taken from: https://www. un. org/en/genocideprevention/documents/U},
  number={20Strategy},
  year={2019}
}

@article{saltelli2002sensitivity,
  title={Sensitivity analysis for importance assessment},
  author={Saltelli, Andrea},
  journal={Risk analysis},
  volume={22},
  number={3},
  pages={579--590},
  year={2002},
  publisher={Wiley Online Library}
}

@article{holzinger2020measuring,
  title={Measuring the quality of explanations: the system causability scale~({SCS})},
  author={Holzinger, Andreas and Carrington, Andr{\'e} and M{\"u}ller, Heimo},
  journal={KI-K{\"u}nstliche Intelligenz},
  pages={1--6},
  year={2020},
  publisher={Springer}
}

@inproceedings{zaidan2007using,
  title={Using annotator rationales to improve machine learning for text categorization},
  author={Zaidan, Omar and Eisner, Jason and Piatko, Christine},
  booktitle={Prof. of Human language technologies 2007: The conference of the North American chapter of ACL},
  pages={260--267},
  year={2007}
}

@article{herrewijnenmachine,
  title={Machine-annotated Rationales: Faithfully Explaining Text Classification},
  author={Herrewijnen, Elize and Nguyen, Dong and Mense, Jelte and Bex, Floris}
}

@article{deyoung2019eraser,
  title={{ERASER}: A benchmark to evaluate rationalized nlp models},
  author={DeYoung, Jay and Jain, Sarthak and Rajani, Nazneen Fatema and Lehman, Eric and Xiong, Caiming and Socher, Richard and Wallace, Byron C},
  journal={arXiv preprint arXiv:1911.03429},
  year={2019}
}

@article{mathew2020hatexplain,
  title={HateXplain: A Benchmark Dataset for Explainable Hate Speech Detection},
  author={Mathew, Binny and Saha, Punyajoy and Yimam, Seid Muhie and Biemann, Chris and Goyal, Pawan and Mukherjee, Animesh},
  journal={arXiv preprint arXiv:2012.10289},
  year={2020}
}

@article{romim2020hate,
  title={Hate Speech detection in the Bengali language: A dataset and its baseline evaluation},
  author={Romim, Nauros and Ahmed, Mosahed and Talukder, Hriteshwar and Islam, Md Saiful},
  journal={arXiv preprint arXiv:2012.09686},
  year={2020}
}

\section*{Appendix}\label{sec:supplements}
Here, we provide more detail about baseline ML/DNN and transformer models. Further, to foster reproducibility, we make available the source codes, data, and interactive notebooks\footnote{{\url{https://github.com/rezacsedu/DeepHateExplainer}}}. This repository will be updated with more reproducible resources, e.g., models, notebooks in the coming weeks.

\subsection*{Training details for DNN baseline models}
The architectural parameters used to train a vanilla CNN, Bi-LSTM, and Conv-LSTM models are listed in \Cref{table:params_LSTM}, \Cref{table:params_BiLSTM}, and \Cref{table:params_CLSTM}, respectively. The BiLSTM model is trained for 500 epochs. The idea is to observe how learning unfolds for each model and how the learning behaviour differs with bidirectional LSTM layers. The placement of bidirectional LSTM layers will create two copies of the hidden layer, one fit in the input sequences as is it is, while the second one on a reversed copy of the input sequence. This will make sure that the TimeDistributed layer\footnote{\url{https://keras.io/api/layers/recurrent_layers/time_distributed/}} receives 100~(or 300) timesteps of 32 outputs, instead of 10 timesteps of 64~(32 units + 32 units) outputs. That is, the first hidden layer will have 100 memory units, while the output layer will be a fully connected layer that outputs one value per timestep. The softmax activation function is used on the output to predict the types of hate. In other words, the output values from both BiLSTM layers will be concatenated, be fed into a fully connected softmax layer for the classification.

\begin{table}[h]
    \centering
    \caption{Parameters for CNN model}
    \label{table:params_LSTM}
    \scriptsize
    \vspace{-4mm}
    \begin{tabular}{l|l}
        \hline
        \textbf{Parameter name} & \textbf{Parameter value}  \\
        \hline
        Embedding dimension & 100, 200, 300 \\
        \hline
        Batch size & 32, 64\\
        \hline
        CNN layer 1 & 64, 128  \\
        \hline
        CNN layer 2 & 32, 64  \\
        \hline
        Pooling size & 2, 3  \\
        \hline
        Dense layer 1 & 128, 256  \\
        \hline
        Dense layer 2 & 256, 512  \\
        \hline
        Dropout & 0.2, 0.3  \\
        \hline
        Gaussian noise & 0.1, 0.2, 0.3, 0.5  \\
        \hline
        Learning rate & 0.001, 0.01, 0.1  \\
        \hline
    \end{tabular}
    \vspace{-2mm}
\end{table}

\begin{table}[h]
    \centering
    \caption{Parameters for Bi-LSTM model}
    \label{table:params_BiLSTM}
    \scriptsize
    \vspace{-4mm}
    \begin{tabular}{l|l}
        \hline
        \textbf{Parameter name} & \textbf{Parameter value}  \\
        \hline
        Embedding dimension & 100, 200, 300  \\
        \hline
        Batch size & 32, 64\\
        \hline
        Bidirectional LSTM layer 1 & 32, 64  \\
        \hline
        Bidirectional LSTM layer 2 & 32, 64  \\
        \hline
        Dense layer 1 & 128, 256  \\
        \hline
        Dense layer 2 & 256, 512  \\
        \hline
        Dropout & 0.2, 0.3  \\
        \hline
        Gaussian noise & 0.1, 0.2, 0.3, 0.5  \\
        \hline
        Learning rate & 0.001, 0.01, 0.1  \\
        \hline
    \end{tabular}
    \vspace{-2mm}
\end{table}

\begin{table}[h]
    \centering
    \caption{Parameters for Conv-LSTM model}
    \label{table:params_CLSTM}
    \scriptsize
    \vspace{-4mm}
    \begin{tabular}{l|l}
        \hline
        \textbf{Parameter name} & \textbf{Parameter value}  \\
        \hline
        Embedding dimension & 100, 200, 300  \\
        \hline
        Batch size & 32, 64\\
        \hline
        CNN layer 1 & 64, 128  \\
        \hline
        CNN layer 2 & 32, 64  \\
        \hline
        Pooling size & 2, 3  \\
        \hline
        LSTM layer 1 & 32, 64  \\
        \hline
        LSTM layer 2 & 32, 64  \\
        \hline
        Dense layer 1 & 128, 256  \\
        \hline
        Dense layer 2 & 256, 512  \\
        \hline
        Dropout & 0.2, 0.3  \\
        \hline
        Gaussian noise & 0.1, 0.2, 0.3, 0.5  \\
        \hline
        Learning rate & 0.001, 0.01, 0.1  \\
        \hline
    \end{tabular}
\end{table}

During the training of Conv-LSTM, an LSTM layer treats an input feature space of $100 \times 300$ and its embedded feature vector dimension as timesteps, which generates 100 hidden units per timestep. Once the embedding layer passes an input feature space $100 \times 300$ into a convolutional layer, the input is padded such that the output has the same length as the original input. Then the output of each convolutional layer is passed to the dropout~(or Gaussian noise) layer to regularize learning to avoid overfitting. This involves the input feature space into a $100 \times 100$ representation, which is then further down-sampled by three different 1D max-pooling layers, each having a pool size of 4 along with the word dimension, each producing an output of shape $25 \times 100$, where each of 25 dimensions can be considered as \emph{extracted features}. Each max-pooling layer follows to \emph{flatten} the output space by taking the highest value in each timestep dimension, which produces a $1 \times 100$ vector that forces words that are highly indicative of interest. These vectors are then fed into a fully connected softmax layer to predict the probability distribution over the hate classes.

\subsection*{Training details for ML baseline models}
\begin{figure}[h]
    \centering
    \includegraphics[scale=0.6]{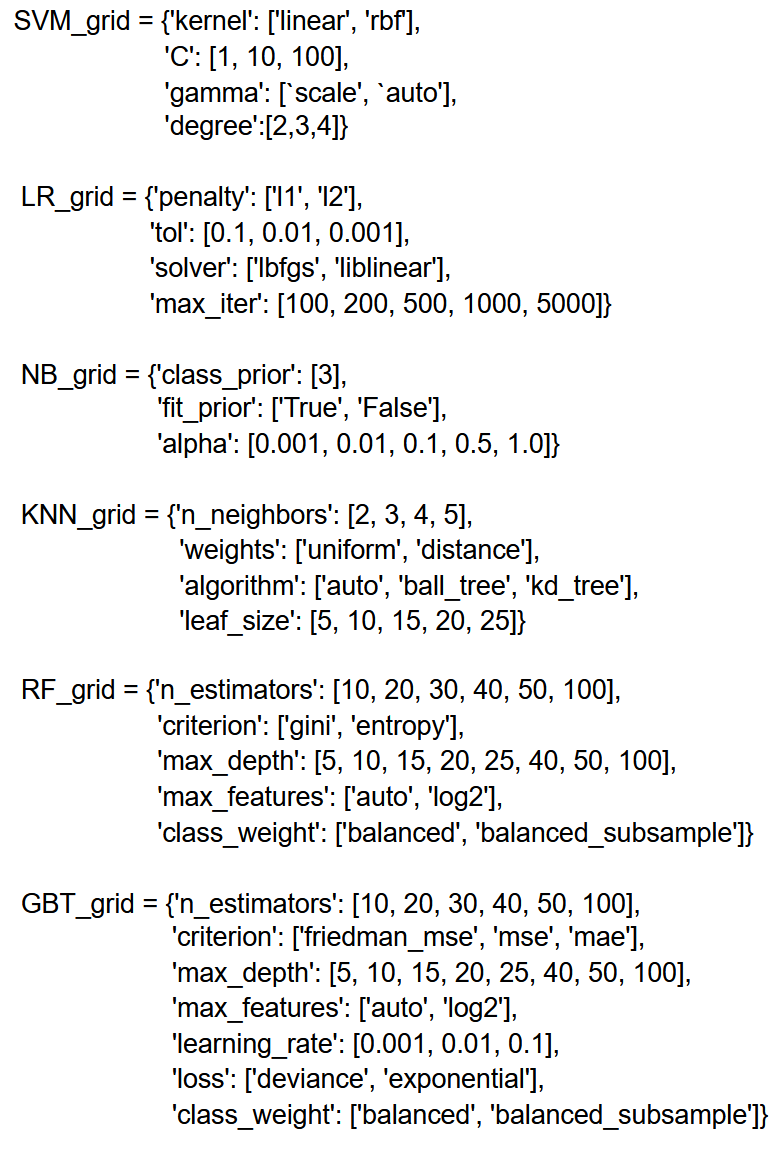}
    \caption{Param grids for ML base line models}
    \label{fig:linear_ML_params}
\end{figure}

We train LR, SVM, KNN, NB, RF, and GBT ML baselines models using the scikit-learn library. We apply both character n-grams and word uni-grams with TF-IDF weighting. The best hyperparameters are produced through random and with 5-fold cross-validation tests. More specifically, Fig. \ref{fig:linear_ML_params} listed the hyperparameters considered in a random search setting. 

\subsection*{Classification results}
We enlist evaluation results of each BERT variant and ensemble of top models on the held-out test set, where the following class encoding to interpret the class-specific classification: i) \emph{personal hate}: class 0, ii) \emph{political hate}: class 1, iii) \emph{religious hate}: class 2, and iv) \emph{geopolitical hate}: class 3. We provide class-wise classification result for each BERT variant, while the same based on the ensemble prediction is shown in \Cref{table:class_specific_result_v2}, covering each hate category. 

\begin{figure}[h]
    \centering
    \includegraphics[width=0.69\linewidth]{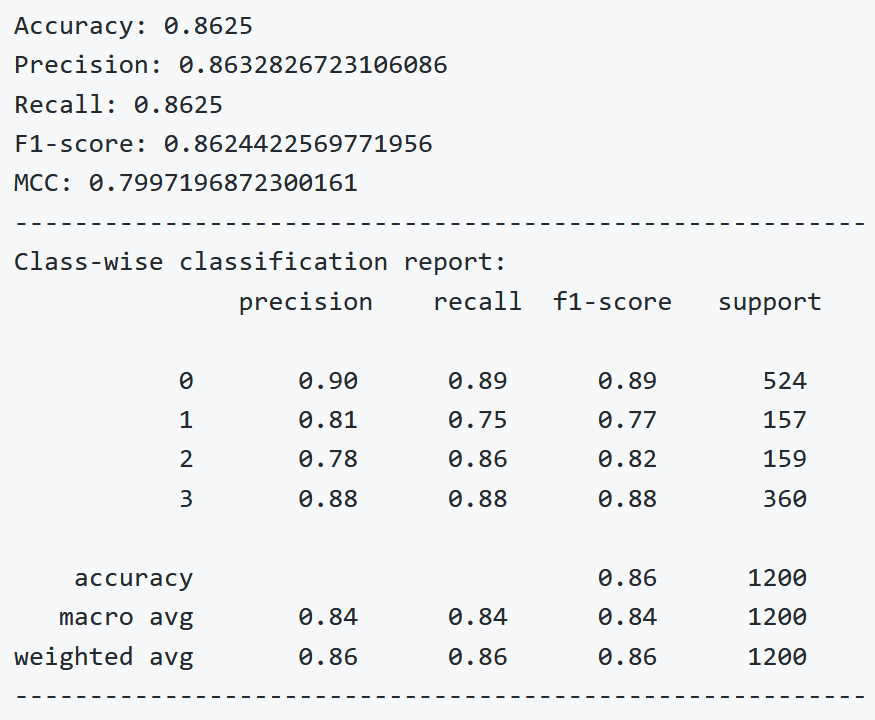}
    \label{fig:1}
    \caption{Class-wise classification results based on Bangla-BERT model}
\end{figure}

\begin{figure}[h]
    \centering
    \includegraphics[width=0.69\linewidth]{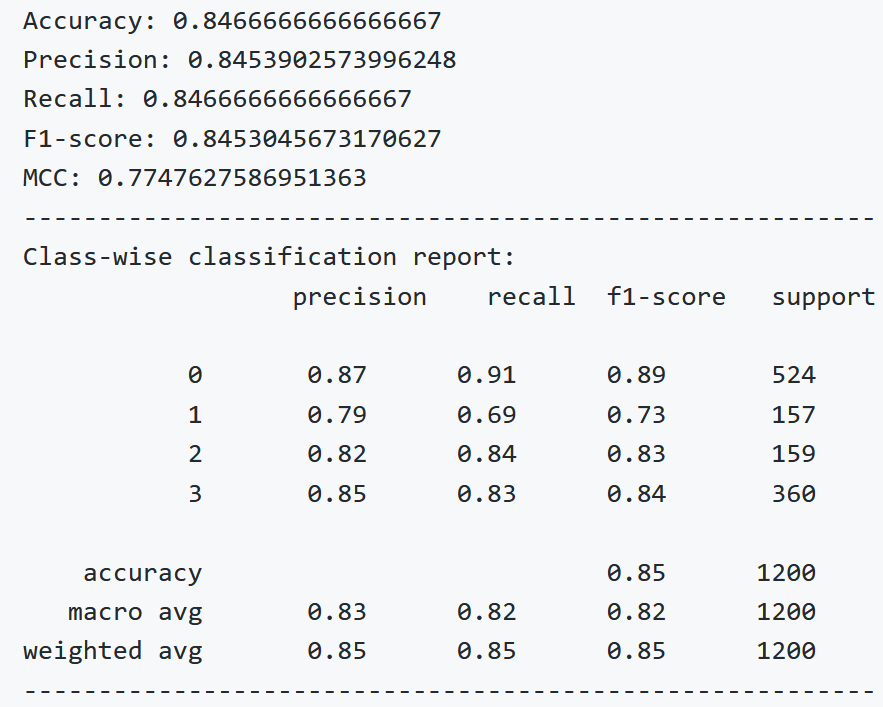}
    \label{fig:2}
    \caption{Class-wise classification results based on BERT-base-multilingual-cased model}
\end{figure}

\begin{figure}[h]
    \centering
    \includegraphics[width=0.69\linewidth]{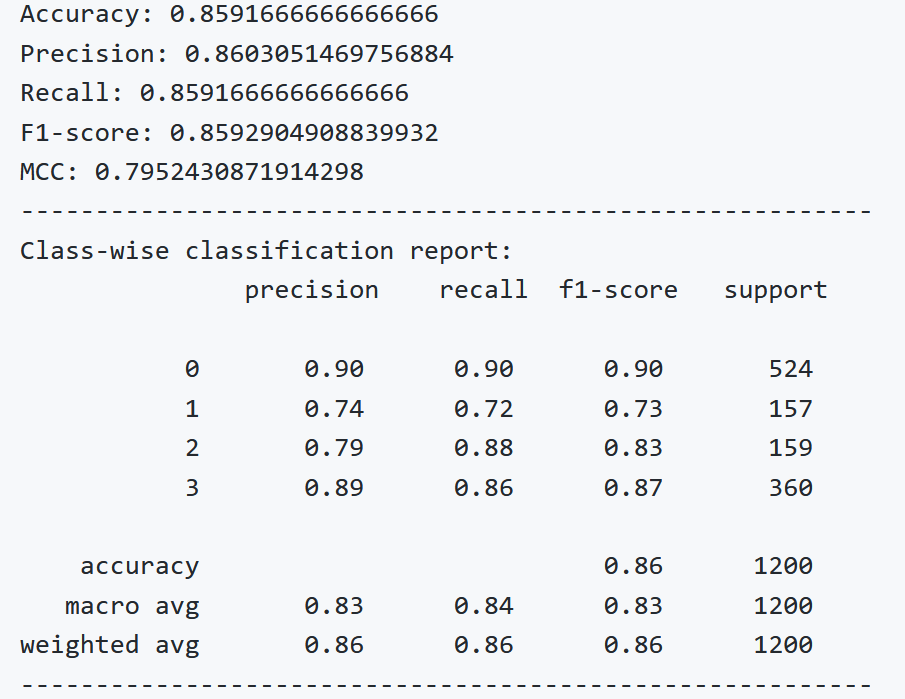}
    \label{fig:3}
    \caption{Class-wise classification results based on BERT-base-multilingual-uncased model}
\end{figure}

\begin{figure}[h]
    \centering
    \includegraphics[width=0.69\linewidth]{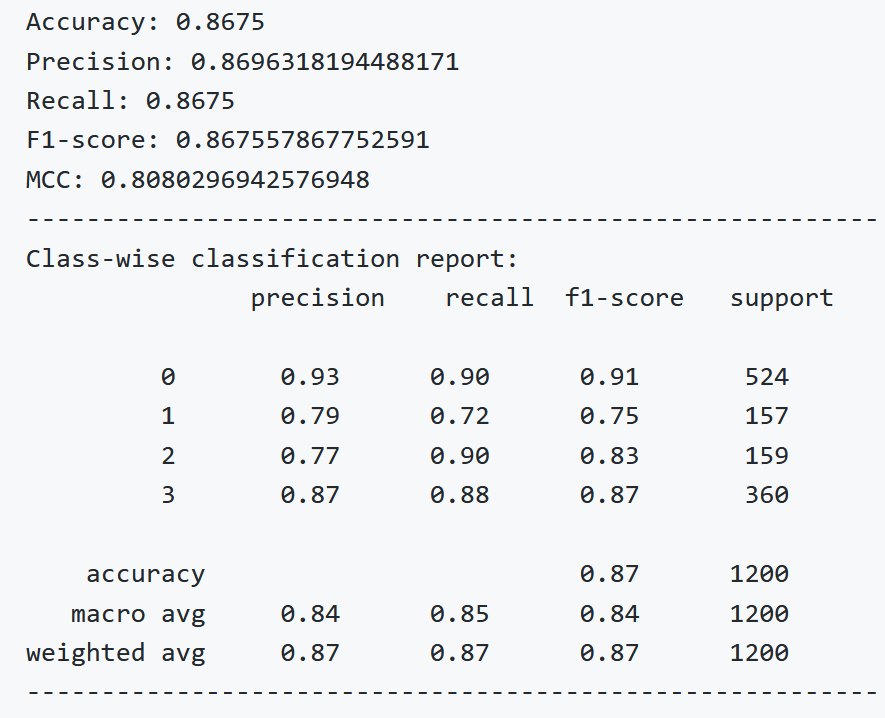}
    \label{fig:4}
    \caption{Class-wise classification results based on XLM-RoBERTa model}
\end{figure}

\begin{figure}[h]
    \centering
    \includegraphics[width=0.9\linewidth]{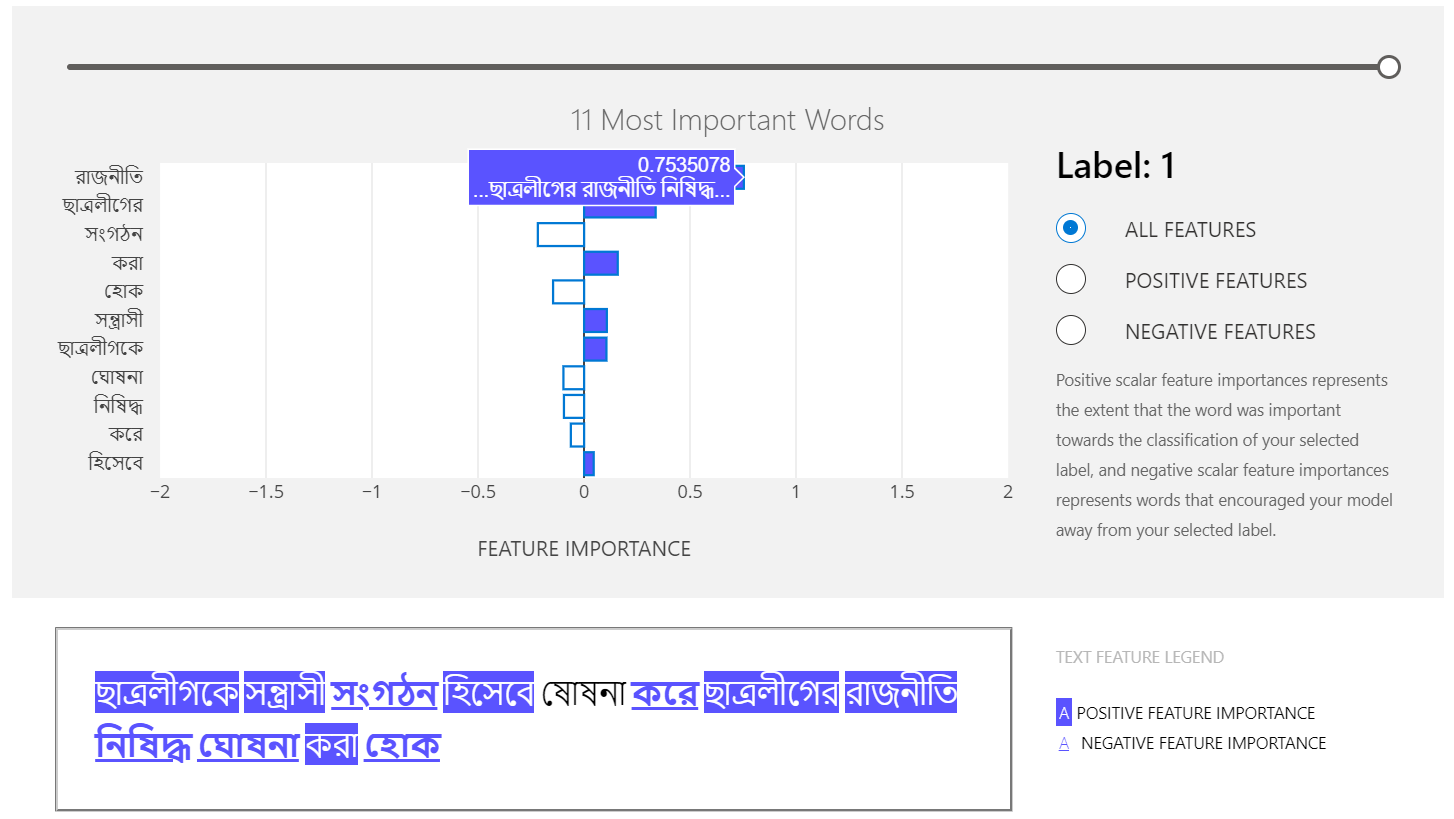}
    \caption{Example-1: identification of political hate, showing most relevant terms}
    \label{fig:example_1}
\end{figure}

\begin{figure}[h]
    \centering
    \includegraphics[width=0.9\linewidth]{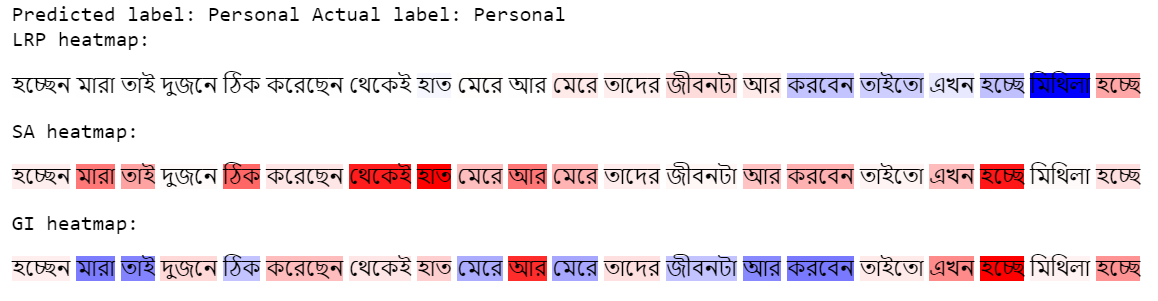}
    \caption{Example-2: identification of personal hate, showing most relevant terms}
    \label{fig:hell2}
\end{figure}

\subsection*{Explanations}
We provide two examples that highlight important terms a DNN model puts more attention to. The example\footnote{ \url{https://github.com/rezacsedu/DeepHateExplainer/blob/main/notebooks/Example_interpret_text.ipynb}} in \cref{fig:example_1}, shows positive feature importances represent the extent that the word was important towards the classification of the selected label, while negative feature importances represents words that encouraged the model away from the selected label. Either positive, negative or both positive and negative features can be selected, outlining their relative importance, while \cref{fig:example_2} shows an example\footnote{\url{https://github.com/rezacsedu/DeepHateExplainer/blob/main/notebooks/LRP_BiLSTM_FastText_Embbeddings_4_Class.ipynb}} detection of personal hate based on using SA, LRP, and integrated gradients~(GI). LRP accurately highlights~(deep blue) most relevant words\includegraphics[width=0.045\textwidth,height=4mm]{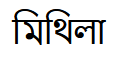}~(Bangladeshi actress Mithila), which signify a personal hate. 

\end{document}